\documentclass[11pt]{article}

\usepackage[margin=1in]{geometry}
\usepackage{amsmath,amssymb,amsthm,mathtools}
\usepackage{microtype}
\usepackage{booktabs}
\usepackage{longtable}
\usepackage{array}
\usepackage{multirow}
\usepackage{graphicx}
\usepackage{xcolor}
\usepackage[numbers,sort&compress]{natbib}
\usepackage[hidelinks]{hyperref}
\usepackage{cleveref}
\usepackage{enumitem}
\usepackage{listings}
\usepackage{tcolorbox}
\usepackage{authblk}
\usepackage{tikz}
\usetikzlibrary{arrows.meta,positioning,fit,backgrounds,calc,shapes.geometric}
\definecolor{cGreen}{HTML}{2E7D32}\definecolor{cGreenBg}{HTML}{E8F5E9}
\definecolor{cBlue}{HTML}{1565C0} \definecolor{cBlueBg}{HTML}{E7F0FA}
\definecolor{cAmber}{HTML}{B26A00}\definecolor{cAmberBg}{HTML}{FDF0DA}
\definecolor{cRed}{HTML}{C0392B}  \definecolor{cRedBg}{HTML}{FBEAE8}
\definecolor{cGray}{HTML}{5F6B7A} \definecolor{cGrayBg}{HTML}{EEF1F4}
\definecolor{cInk}{HTML}{22303C}

\tikzset{
  apbox/.style   ={rounded corners=3pt, draw=cGray, line width=0.7pt, fill=white,
                   align=center, text=cInk, inner sep=3.5pt, font=\scriptsize},
  apstage/.style ={apbox, draw=cBlue, fill=cBlueBg, line width=0.8pt, minimum height=10mm},
  apcheck/.style ={apbox, draw=cAmber, fill=cAmberBg, line width=0.8pt},
  apgood/.style  ={apbox, draw=cGreen, fill=cGreenBg, line width=0.8pt},
  apquiet/.style ={apbox, draw=cGray, fill=cGrayBg},
  apar/.style    ={-{Stealth[length=4.5pt]}, line width=0.9pt, draw=cGray},
  aparg/.style   ={apar, draw=cGreen},
  apara/.style   ={apar, draw=cAmber},
  aparb/.style   ={apar, draw=cRed, dashed},
  aplab/.style   ={font=\tiny, text=cGray, inner sep=1.5pt},
  aplabg/.style  ={aplab, text=cGreen},
  aplabr/.style  ={aplab, text=cRed},
}
\newcommand{\aploopunder}[4][6mm]{%
  \draw[aparb] #2 -- ++(0,-#1) -- node[aplabr, below=0.5pt] {#4}
    ($#3+(0,-#1)$) -- #3;}
\newcommand{\aploopover}[4][5mm]{%
  \draw[aparb] #2 -- ++(0,#1) -- node[aplabr, above=0.5pt] {#4}
    ($#3+(0,#1)$) -- #3;}

\theoremstyle{plain}

\theoremstyle{definition}

\theoremstyle{remark}

\newcommand{\Causalean}{\textsc{Causalean}}
\newcommand{\CausalSmith}{\textsc{CausalSmith}}
\newcommand{\lean}{\textsf{Lean~4}}
\newcommand{\code}[1]{\texttt{#1}}

\newcommand{\ub}{\_\allowbreak}

\newcommand{\statusgated}{\textsf{gated}}
\newcommand{\statuscited}{\textsf{cited}}

\lstdefinelanguage{lean}{
  morekeywords={theorem,lemma,def,structure,class,instance,axiom,sorry,by,
    intro,intros,apply,exact,simp,rw,have,show,fun,match,with,end,namespace,
    variable,variables,open,import,example},
  sensitive=true,
  morecomment=[l]{--},
  morecomment=[s]{/-}{-/},
  morestring=[b]",
}
\title{\CausalSmith: A Formally Grounded, Self-Improving Agentic\\
Framework for Automated Research in Causal Inference}
\author[]{Jiyuan Tan \quad \quad Vasilis Syrgkanis}
\date{}

\begin{document}
\maketitle

\begin{abstract}
Automating theoretical research requires generating candidate results and evaluating them reliably. Models keep getting better at the first, while the second remains hard. A common approach asks one large language model (LLM) to review what another produced, yet such reviewers are empirically unreliable: they may accept fabricated papers and catch the fabrication at close to chance rates~\citep{badscientist2025}. We present \textsc{CausalSmith}, a framework for automated theoretical research in causal inference built on the Lean proof assistant, where a proof is checked by a program rather than read by a referee. \textsc{CausalSmith} rests on \textsc{Causalean}, a foundational Lean library for causal inference holding 8,179 machine-checked definitions and theorems, developed with language-model assistance under human design and review. Around it we build a self-improving agentic pipeline that selects research topics, proposes results, formalizes statements, constructs proofs, and presents the resulting artifacts for human inspection. Moreover, the pipeline pairs Lean verification with a statement audit that compares each formal theorem against the informal claim behind it. We evaluate the system using artifacts produced by completed autonomous research runs. The source code, formal library, and run records are available at \url{https://github.com/Jiyuan-Tan/CausalSmith}.

\end{abstract}

\section{Introduction}
\label{sec:intro}

Language models can now generate research artifacts---conjectures, proofs,
experiments, and complete papers---more rapidly than they can be evaluated. In
empirical fields, this primarily increases the burden of review. In theoretical work,
it presents a more direct risk: an incorrect theorem may be indistinguishable from a
correct one until it is verified by an expert, but expert verification is
slow and costly. Many systems for automated mathematical and theoretical research
address this bottleneck by delegating evaluation to another language model: one model proposes
and another reviews.

Empirical evaluations already show the defects of LLM reviewers. Independent evaluations of the
AI-Scientist systems~\citep{aiscientist2025} report hallucinated numbers, frequent
coding failures, and little novelty~\citep{aiscientist_eval2025}.
\citet{badscientist2025} further show that LLM reviewers
accept deliberately fabricated papers up to $82\%$ of the time and detect the
fabrication at near-chance rates. These findings make LLM review an unreliable evaluator of correctness.

Formal verification offers a different basis for evaluation. A theorem written in
\lean{} is checked by a program rather than read by a referee. If \lean{} accepts the
proof, then every step of the argument has been verified against the axioms, so a gap or a mistake in the reasoning cannot survive. Grounding a system in \lean{} takes the correctness of proofs
out of a model's hands entirely. However, two obstacles stand between that guarantee and an
automated system that can use it: cost and faithfulness.

The first obstacle is cost. Formalizing a result from first principles is slow work,
so a system that proves causal-inference theorems needs a library to start from:
identification theorems, estimators, and asymptotic results already stated and
proved in the literature. Without one, every run rebuilds standard theory before it can reach anything
new.

The second obstacle is faithfulness. \lean{} checks the proof, but it says nothing
about whether the theorem is the one the system meant to state. As a result, the natural-language statement may not match the theorem actually proved. Recent work
documents this gap from several angles~\citep{sorries2026, signalcoverage2026,
beyondkernel2026}, and we saw it in our own runs. In one, an agent turned the hard
lemmas into \code{axiom}s and produced \lean{} code that compiled with axioms
standing in for proofs. In another, the model altered a hypothesis, so the \lean{}
code proved a weaker statement than the paper claimed. Both compile, and neither is a
proof of what was claimed. Our pipeline guards against this with a fine-grained audit step
(\Cref{sec:pipeline-audit}) that compares each formal statement with the claim it is
meant to express. \lean{} checks the proofs the pipeline produces, and the audit
checks that the statements they establish are the ones the paper reports.

We instantiate this approach in causal inference through \CausalSmith{}, which has
two components. The first is \Causalean{}, a \lean{} library covering the
causal-inference toolkit, which gives agents verified pieces to build on. It was
itself built with LLM assistance: humans chose what to formalize and how
to state it, agents drafted the definitions, proofs, and docstrings under those
decisions, and a human then reviewed the resulting \lean{} statements. The second
component is a self-improving agentic pipeline that selects or accepts a research
topic, proposes a result, formalizes it, proves it, and presents it. It audits one
statement at a time rather than judging a finished paper once, and it promotes
reusable lemmas into \Causalean{} as they are proved and reviewed.

\paragraph{Contributions.}
\begin{itemize}[leftmargin=1.4em,itemsep=1pt,topsep=2pt]
  \item \Causalean{}, a broad \lean{} formalization of causal inference spanning
  graphical and structural causal models, potential outcomes and identification,
  panel methods, experimentation, estimation, and statistical theory. It comprises
  $8{,}179$ machine-checked declarations, meaning named definitions and theorems that
  \lean{} has verified. It was written with language-model assistance under human
  design, and it provides a retrieval interface for agents (\Cref{sec:library}).
  \item An end-to-end research pipeline whose Discovery stage can select its
  own research topic and propose a causal-inference result, which the pipeline then
  formalizes, proves, and presents. It also contains a library feedback loop that
  adds reusable lemmas and theorems proved during a run back into \Causalean{}
  (\Cref{sec:pipeline,sec:pipeline-audit}).
  \item An evaluation drawn from $144$ recorded runs: a catalog of machine-checked
  results and a headline result found by the system, which closes a gap in \citet{zeng2024discrete}
  (\Cref{sec:eval}).
\end{itemize}

\begin{figure*}[!htb]
  \centering
  \begin{tikzpicture}[
      font=\footnotesize, node distance=18mm,
      stage/.style={rounded corners=3pt, draw=cBlue, line width=0.8pt, fill=cBlueBg,
                    minimum height=13mm, minimum width=27mm, align=center, text=cInk},
      gate/.style={rounded corners=2pt, draw=cAmber, line width=0.7pt, fill=cAmberBg,
                   minimum height=6mm, inner sep=3pt, align=center, font=\scriptsize,
                   text=cInk},
      lib/.style={rounded corners=3pt, draw=cGreen, line width=0.9pt, fill=cGreenBg,
                  minimum height=12mm, minimum width=34mm, align=center, text=cInk},
      bank/.style={rounded corners=3pt, draw=cAmber, line width=0.9pt, fill=cAmberBg,
                   minimum height=9mm, align=center, text=cInk, inner sep=4pt},
      fl/.style={-{Stealth[length=5pt]}, line width=0.9pt, draw=cGray},
      dfl/.style={-{Stealth[length=5pt]}, line width=0.9pt, draw=cGreen, dashed},
      afl/.style={-{Stealth[length=5pt]}, line width=0.9pt, draw=cAmber, dashed},
    ]
    \node[stage] (disc) {\textbf{Discover}\\[1pt]{\scriptsize propose $\to$ solve}};
    \node[stage, right=of disc] (form) {\textbf{Formalize}\\[1pt]{\scriptsize plan $\to$ scaffold}};
    \node[stage, right=of form] (prove) {\textbf{Prove}\\[1pt]{\scriptsize fill $\to$ review}\\[1pt]{\scriptsize\itshape type-checked}};
    \node[stage, right=of prove] (pres) {\textbf{Present}\\[1pt]{\scriptsize assemble \& present}};

    \draw[fl] (disc) -- node[gate, above=1.5mm] {novelty\\[-1pt]+ soundness} (form);
    \draw[fl] (form) -- node[gate, above=1.5mm] {statement\\[-1pt]match} (prove);
    \draw[fl] (prove) -- node[gate, above=1.5mm] {convergence\\[-1pt]review} (pres);

    \node[lib, below=12mm of $(form)!0.5!(prove)$] (lib)
      {\Causalean{} \;\textit{\scriptsize verified library}};
    \draw[dfl] (lib.north west) to[out=120,in=-90]
      node[left=1pt, font=\scriptsize, text=cGreen, pos=0.6] {reuse} (form.south);
    \draw[dfl] (prove.south) to[out=-90,in=60]
      node[right=1pt, font=\scriptsize, text=cGreen, pos=0.42, align=center] {study\\mode} (lib.north east);
    \node[bank, minimum width=15.5cm, below=8mm of lib] (bank)
      {run record \;\textit{\scriptsize prior proposals \& outcomes}};
    \draw[afl] (bank.north -| disc.south) --
      node[left=1pt, font=\scriptsize, text=cAmber, pos=0.5] {screen} (disc.south);
    \draw[afl] (pres.south) --
      node[right=1pt, font=\scriptsize, text=cAmber, pos=0.5] {record}
      (bank.north -| pres.south);
  \end{tikzpicture}
  \caption{The \CausalSmith{} pipeline. A claim moves left to right through discovery,
  formalization, proof, and presentation, clearing a review between stages: novelty and
  mathematical soundness, statement matching, and a final dual-model convergence
  review of the frozen graph. Two feedback channels grow the system
  (\Cref{sec:pipeline-study}): study mode proves a load-bearing lemma \Causalean{}
  lacks and promotes it back into the library, and every run---accepted, downgraded,
  or failed---enters the run record that screens the next proposal.
  \Cref{app:pipeline-spec,app:pipeline-proof-loop,app:pipeline-paper,app:study}
  give the operational detail.}
  \label{fig:pipeline}
\end{figure*}
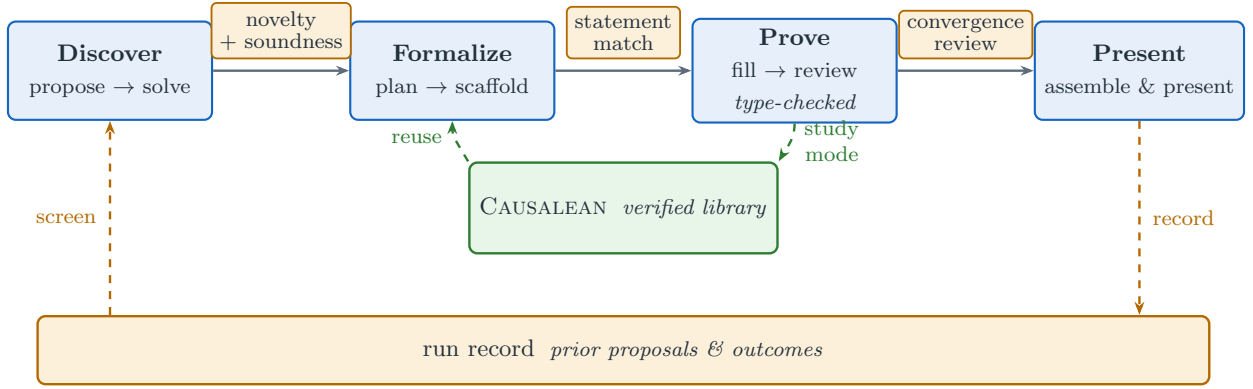

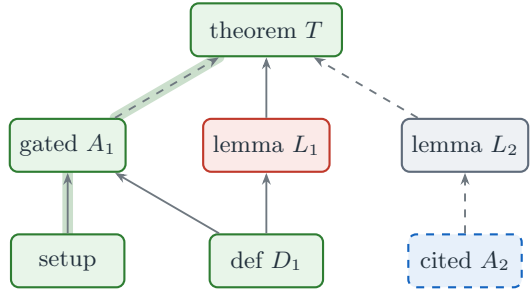
\begin{figure}[t]
  \centering
  \begin{tikzpicture}[
      node distance=8mm and 11mm, font=\footnotesize,
      nd/.style={rounded corners=3pt, draw, minimum height=7mm, minimum width=15mm,
                 align=center, line width=0.8pt, text=cInk},
      matched/.style={nd, fill=cGreenBg, draw=cGreen},
      drift/.style={nd, fill=cRedBg, draw=cRed},
      unrev/.style={nd, fill=cGrayBg, draw=cGray},
      cited/.style={nd, fill=cBlueBg, draw=cBlue, dashed},
      suse/.style={-{Stealth[length=4.5pt]}, draw=cInk!65, line width=0.7pt},
      puse/.style={-{Stealth[length=4.5pt]}, draw=cInk!65, line width=0.7pt, dashed},
      crit/.style={line width=4pt, draw=cGreen!25, line cap=round},
    ]
    \node[matched] (s)  {setup};
    \node[matched, right=of s] (d1) {def $D_1$};
    \node[cited,  right=of d1] (a2) {cited $A_2$};
    \node[matched, above=of s]  (a1) {gated $A_1$};
    \node[drift,   above=of d1] (l1) {lemma $L_1$};
    \node[unrev,   above=of a2] (l2) {lemma $L_2$};
    \node[matched, above=of l1, minimum width=20mm] (t) {theorem $T$};

    \begin{scope}[on background layer]
      \draw[crit] (s) -- (a1);
      \draw[crit] (a1) -- (t);
    \end{scope}

    \draw[suse] (s)  -- (a1);
    \draw[suse] (d1) -- (l1);
    \draw[suse] (d1) -- (a1);
    \draw[puse] (a2) -- (l2);
    \draw[puse] (a1) -- (t);
    \draw[suse] (l1) -- (t);
    \draw[puse] (l2) -- (t);
  \end{tikzpicture}
  \caption{A dependency graph (schematic). Nodes are statements; fill encodes review status
  (\textcolor{cGreen}{\textbf{matched}}, \textcolor{cRed}{\textbf{drift}},
  \textcolor{cGray}{\textbf{unreviewed}}), and a dashed blue border marks a cited
  node. Solid edges show what a claim says, dashed edges what its proof uses;
  the green underlay is the critical path of must-prove nodes
  the review must clear.}
  \label{fig:graph}
\end{figure}

\paragraph{\CausalSmith{} pipeline.}
The pipeline has four stages, shown in \Cref{fig:pipeline}. Discovery takes a topic
from a researcher or proposes one itself, then solves it in natural language under an
adversarial review of novelty and soundness. We store that solution as a dependency graph: its nodes are the individual statements of the paper---a setup, a
definition, an assumption, a lemma, the headline theorem---and its edges record which
statements a claim uses and which its proof consumes (\Cref{fig:graph}).
Formalization gives every node a \lean{} declaration, reusing what \Causalean{}
already provides, and the statement audit checks node by node that each declaration
says what its node claims. Proof construction fills those declarations against the
compiler and promotes any reusable lemma the library lacked back into \Causalean{}.
Presentation writes the paper from the finished graph, and the run---accepted,
downgraded, or failed---enters the record that screens the next proposal.
\Cref{sec:pipeline} describes the stages in full.

\paragraph{Availability.}
The \Causalean{} library, the research pipeline, and the run record backing
\Cref{sec:eval} are available at \url{https://github.com/Jiyuan-Tan/CausalSmith}. A
companion site at \url{https://jiyuan-tan.github.io/CausalSmith/} provides a browsable
view of the library, including the natural-language statement of each declaration and
the generated write-up of each accepted result. Each write-up carries a proof map of its
own results, a link from every statement to the \lean{} declaration it was audited
against, and a slide deck. Readers can verify, rate, and comment on individual
statements. \Cref{app:site} shows these.

\section{Related Work}
\label{sec:related}

We discuss four lines of work: language models paired with proof assistants, and
whether the statements they produce mean what was intended; agentic systems for
automated research, and the verifiers their output is judged by; language models
applied to causal inference; and formal libraries for statistics and economics.
\Cref{tab:positioning} summarizes where \CausalSmith{} sits among the systems closest
to it.

\begin{table}[t]
  \centering
  \footnotesize
  \setlength{\tabcolsep}{4.5pt}
  \caption{Where \CausalSmith{} sits among neighboring systems. ``Statement
  audit?'' asks whether a system checks that the formal statement means the intended
  claim, beyond type-checking.}
  \label{tab:positioning}
  \begin{tabular}{lccccc}
    \toprule
    System & Proposes & Machine-checked & Statement & Evaluator & Self- \\
           & theorem? & proof? & audit? & soundness & improving? \\
    \midrule
    AI Scientist~\citep{aiscientist2025}      & \checkmark & --- & --- & LLM judge & --- \\
    FunSearch\,/\,AlphaEvolve~\citep{funsearch2024,alphaevolve2025} & partial & \checkmark & --- & task metric & \checkmark \\
    AlphaProof~\citep{alphaproof2025}         & --- & \checkmark & n/a & sound & partial \\
    Causal benchmarks~\citep{cladder2023,corr2cause2024} & --- & --- & --- & fixed labels & --- \\
    Causal agents~\citep{causalcopilot2024}   & --- & --- & --- & none & --- \\
    Lean economics library~\citep{econcslib2026}     & --- & \checkmark & --- & sound & --- \\
    Asymptotic-statistics agents~\citep{asymptoticstats2026} & --- & \checkmark & \checkmark & sound + audit & --- \\
    \textbf{\CausalSmith{} (ours)}            & \checkmark & \checkmark & \checkmark & sound + audit & \checkmark \\
    \bottomrule
  \end{tabular}
\end{table}

\paragraph{Autoformalization, theorem proving, and statement matching.}
Language models have been paired with proof assistants for neural proof
search~\citep{gptf2020,htps2022}, autoformalization~\citep{wu2022autoform,dsp2022},
whole-proof generation and repair~\citep{baldur2023}, retrieval-augmented
proving~\citep{leandojo2023,deepseekprover2025}, and competition
mathematics~\citep{alphageometry2024,alphaproof2025}, and they are measured against
benchmarks of known problems~\citep{minif2f2021,proofnet2023,putnambench2024}.
Closest to us is autoformalization, where a model turns an informal statement into a
formal one. The result is compared against a human-written formal
version: by expert inspection~\citep{wu2022autoform}, by roundtrip and
back-translation checks~\citep{faithfulnessgap2026,roundtrip2026,backtrans2024}, or by
learned alignment scorers~\citep{formalalign2025}. Surface-overlap metrics have been
rejected as evidence of a match~\citep{bleu_disavow2024}. Our setting has no such
reference. The theorem is new, so the run writes its own informal claim and then
formalizes it, leaving both sides of the comparison model output. Studies of this gap
report formalizations with no remaining proof gaps that still fail expert
review~\citep{sorries2026}, statements a proof assistant accepts that are semantically
wrong~\citep{signalcoverage2026,beyondkernel2026}, and vacuity and reward hacking
common enough to benchmark~\citep{formalreward2026}. Our audit therefore
back-translates, inside the discovery loop and localized over a proof-dependency
graph, in the style of Lean blueprints and proof-flow
tools~\citep{leanarchitect2026,proofflow2025}, within the broader shift toward
proof-assistant-grounded reasoning~\citep{formalfrontier2024}.

\paragraph{Automated research agents and verified discovery.}
Agents that carry out end-to-end research~\citep{aiscientist2025} are vulnerable when
an LLM judges their output, whether shown by independent
re-evaluation~\citep{aiscientist_eval2025} or by adversarial
construction~\citep{badscientist2025}. Other work grounds discovery in a hard
verifier: program search against a fitness
function~\citep{funsearch2024,alphaevolve2025}, or a proof against a type checker.
Those fitness functions are fixed in advance, so a statement audit is a step they
never need. ``Co-scientist'' agents fall in between: they propose and refine
scientific claims but validate them empirically~\citep{aicoscientist2025}. The system
closest to ours is a multi-agent pipeline for asymptotic statistics, whose auditor and
reviewer agents already guard against vacuity~\citep{asymptoticstats2026}. It
formalizes known theory, whereas \CausalSmith{} proposes the theorem as well, and
\CausalSmith{} localizes the audit over the dependency graph, so a changed statement
re-opens the affected nodes rather than the whole formalization.

\paragraph{LLMs for causal inference.}
A parallel literature asks whether language models reason causally, treating causal
reasoning as something to test or deploy rather than something to prove; recent
surveys organize it by causal task and intervention level~\citep{ma2025causal}.
Benchmarks probe graphical and correlational
reasoning~\citep{cladder2023,corr2cause2024}, broad and data-grounded question
answering~\citep{calm2024,qrdata2024}, and, closest to our own domain, end-to-end
causal inference on real scientific
studies~\citep{causcibench2025,causalbench_ours2026}. A critical strand reports that
surface accuracy conceals shallow reasoning: models explain causal language while
hallucinating the argument~\citep{gao2023chatgptcausal}, recite memorized facts rather
than reason interventionally~\citep{causalparrots2023}, commit elementary
fallacies~\citep{fallacies2024}, and are flattered by benchmarks solvable through
lookup~\citep{causalreview2024}. A third strand builds agents that do causal analysis
without a correctness guarantee~\citep{kiciman2023causal,matmcd2024,orca2025,%
causalcopilot2024}, including pipelines that map a dataset and question to an
estimate~\citep{causalaiscientist2026} and systems that search for instrumental
variables~\citep{miningcausality2024,ivcoscientist2026}. These share our
propose-and-critique structure but validate statistically or empirically; none
produces a machine-checked causal theorem. Pairing language models with formal
verification appears confined to provers for general
mathematics~\citep{leancopilot2024}, and \CausalSmith{} works at that intersection.

\paragraph{Formal libraries for statistics and economics.}
A \lean{} ecosystem for theoretical statistics has recently emerged: Statlib on the
decision-theoretic foundations of inference~\citep{statlib2026}, StatsMLlib on
concentration, empirical processes, and finite-sample learning
guarantees~\citep{statsmllib2026}, and a statistical-learning-theory
line~\citep{leanslt2026} extending the Rademacher-complexity library that
\Causalean{} adapts~\citep{leanrademacher2025}. A \lean{} library for economics
formalizes established theory in the same spirit~\citep{econcslib2026}. \Causalean{}
overlaps with all of them wherever causal estimation theory needs the same
tools---minimax lower-bound methods, information divergences, empirical-process and
concentration lemmas---and we expect these projects to benefit as they converge on
shared upstream infrastructure. Those libraries
formalize classical theory as the end product, whereas \Causalean{}'s statistical
clusters exist to support theorems about causal estimands, and its core---structural
causal models, do-calculus, graphical and partial
identification, panel estimand characterizations, and design-based inference under
interference---has no counterpart in any of them. \Causalean{} is also built to be
extended rather than finished: each accepted run can promote its reusable lemmas back
into the library for later runs to use.

\section{Background}
\label{sec:prelim}

\paragraph{Causal inference.}
Causal inference concerns the consequences of intervention rather than observation
alone. Two frameworks are widely used: structural causal models (SCMs) and potential
outcomes (PO). An SCM models causal relationships with a directed acyclic graph and
represents interventions with the $\mathrm{do}$-operator; its classical results
include the graphical identification criteria of do-calculus and the ID algorithm.
Potential outcomes instead attach to each unit a family of counterfactual responses,
one per treatment level, and write causal estimands---the average treatment effect,
the effect on the treated, difference-in-differences, the local average treatment
effect---as functionals of their joint distribution. An estimand is identified when it
is a function of the observational distribution alone and partially identified when
the data determine only a set containing it, as with Manski or Balke--Pearl bounds.
\Causalean{} (\Cref{sec:library}) formalizes most classical results under these
frameworks.

\paragraph{\lean{} as a proof checker.}
\lean{} is a proof assistant and programming language in which mathematics is written
out in complete formal detail~\citep{lean42021}. A proposition is a type and a proof of
it is a value of that type, so checking a proof means checking that a value has the
type it claims; this check is called type-checking. The unit being checked is a
declaration, a single named definition or theorem, and we say a declaration
type-checks when it passes. Mathlib, \lean{}'s community mathematics library, supplies
the analysis, probability, and measure theory that our library
builds on~\citep{mathlib2020}.

This makes \lean{} useful as an evaluator. The check is mechanical and indifferent to
how the proof is written. When a proof of a statement $\tau$ passes, $\tau$ follows from its stated assumptions
together with the axioms the proof invokes, and \lean{} reports those axioms on
request. We call this property proof soundness, but it says nothing about whether
$\tau$ is the proposition the researcher intended to prove, which
\Cref{sec:pipeline-audit} takes up.

Three things pass the type check while leaving the mathematics incomplete, and all
three appear later in the paper. Writing \code{sorry} in place of a proof tells \lean{}
to grant the statement, the formal counterpart of a \code{TODO}. Declaring an
\code{axiom} asserts a statement without proof. And a statement can be trivial: a
definition that unfolds to \code{True} is satisfied by everything, so a theorem about
it holds vacuously. \lean{} reports the first two on request---\code{sorry} raises a
warning, and \code{\#print axioms} lists what a theorem ultimately rests on---while the
third is invisible to \lean{} and has to be caught by reading the statement.

\section{The \Causalean{} Library}
\label{sec:library}

The cost of formalizing results from first principles limits automated discovery.
\Causalean{} is a \lean{} library for causal inference whose verified results agents
can reuse and compose across runs.

\subsection{Design and scope}
\label{sec:library-design}

\Causalean{} holds mathematics that belongs to no single paper: the definitions and theorems any
causal result might reuse. One-off lemmas of a particular paper live instead in the
pipeline package. A \CausalSmith{} run reads from the library freely and writes back to it only by
promoting a reusable lemma, which the pipeline does on its own once the lemma clears
the checks of \Cref{sec:pipeline-study}, and an automated check enforces that boundary
on every build.

The library was built with LLM assistance, following the
division of labor of \Cref{sec:intro}. Humans set the scope, design the structure of the library, choose which results to
formalize, and fix the definitions that later theorems are stated against; agents
draft the statements, proofs, and docstrings. Nothing
enters the library until it passes the \lean{} type check and clears the
axiom check
(\Cref{sec:library-axiom-checks}). A human then reviews that the formal statement
matches the intended claim and that it is stated in a standard form.

The compiled environment index records $8{,}179$ declarations: $5{,}375$ theorems,
$2{,}372$ definitions, and $432$ structures, instances, and inductive types,
across $1{,}070$ files and roughly $302{,}000$ lines (\Cref{tab:coverage}).\footnote{These
repository statistics are a snapshot. The totals may increase because the
\CausalSmith{} pipeline runs continuously and can promote newly proved, reusable
declarations to \Causalean{}.} The
declarations are grouped into ten clusters, from graphical foundations to asymptotic
statistics.

\begin{table}[t]
  \centering
  \small
  \caption{\Causalean{} coverage by cluster. All figures are read from the compiled
  environment index (\code{lake exe library\_index}); the per-cluster kind columns
  sum to the grand total. ``s/c/i'' abbreviates the remaining declaration
  kinds---structures and classes, inductives, and instances.}
  \label{tab:coverage}
  \begin{tabular}{lrrrrr}
    \toprule
    Cluster & Files & Lines & Defs & Thms/Lem. & s/c/i \\
    \midrule
    Stat            & 254 & 67{,}065 & 358 & 1{,}229 &  42 \\
    Estimation      & 168 & 55{,}788 & 300 &    581 & 100 \\
    PO              & 138 & 42{,}587 & 594 &    929 &  79 \\
    Mathlib (local) & 157 & 39{,}550 & 306 &    894 &  36 \\
    SCM             &  88 & 36{,}901 & 209 &    548 &  66 \\
    Panel           &  68 & 19{,}176 & 260 &    423 &  45 \\
    Experimentation & 103 & 17{,}212 & 175 &    363 &  12 \\
    Graph           &  21 & 10{,}948 &  70 &    203 &  28 \\
    ML              &  49 &  6{,}457 &  56 &     95 &  13 \\
    Discovery       &  24 &  6{,}441 &  44 &    110 &  11 \\
    \midrule
    \textbf{Total}  & \textbf{1{,}070} & \textbf{302{,}125} & \multicolumn{3}{c}{\textbf{$8{,}179$ declarations}} \\
    \bottomrule
  \end{tabular}
\end{table}

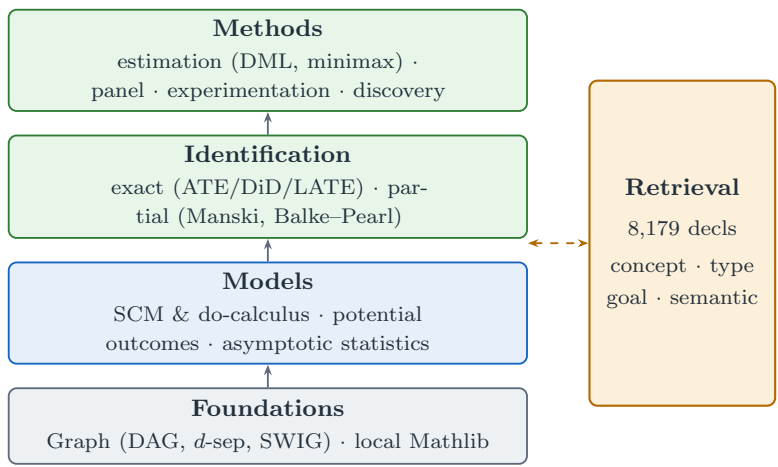
\begin{figure}[t]
  \centering
  \begin{tikzpicture}[
      font=\footnotesize, node distance=3mm,
      lay/.style={rounded corners=3pt, draw, line width=0.7pt, minimum height=10mm,
                  align=center, text width=66mm, text=cInk},
      found/.style={lay, fill=cGrayBg,  draw=cGray},
      core/.style={lay,  fill=cBlueBg,  draw=cBlue},
      app/.style={lay,   fill=cGreenBg, draw=cGreen},
      idx/.style={rounded corners=3pt, draw=cAmber, line width=0.8pt, fill=cAmberBg,
                  minimum height=43mm, text width=22mm, align=center, text=cInk},
      up/.style={-{Stealth[length=4pt]}, line width=0.7pt, draw=cGray},
    ]
    \node[found] (found) {\textbf{Foundations}\\[1pt]{\scriptsize Graph (DAG, $d$-sep, SWIG) $\cdot$ local Mathlib}};
    \node[core,  above=of found] (core)  {\textbf{Models}\\[1pt]{\scriptsize SCM \& do-calculus $\cdot$ potential outcomes $\cdot$ asymptotic statistics}};
    \node[app,   above=of core]  (idpar) {\textbf{Identification}\\[1pt]{\scriptsize exact (ATE/DiD/LATE) $\cdot$ partial (Manski, Balke--Pearl)}};
    \node[app,   above=of idpar] (est)   {\textbf{Methods}\\[1pt]{\scriptsize estimation (DML, minimax) $\cdot$ panel $\cdot$ experimentation $\cdot$ discovery}};

    \node[idx, anchor=west] at ($(est.east)!0.5!(found.east) + (8mm,0)$) (idx)
      {\textbf{Retrieval}\\[4pt]{\scriptsize $8{,}179$ decls}\\[3pt]{\scriptsize concept $\cdot$ type}\\[1pt]{\scriptsize goal $\cdot$ semantic}};

    \begin{scope}[on background layer]
      \draw[up] (found.north) -- (core.south);
      \draw[up] (core.north)  -- (idpar.south);
      \draw[up] (idpar.north) -- (est.south);
    \end{scope}
    \draw[{Stealth[length=4pt]}-{Stealth[length=4pt]}, dashed, line width=0.7pt, draw=cAmber]
      ($(est.east)!0.5!(found.east)$) -- (idx.west);
  \end{tikzpicture}
  \caption{\Causalean{} in layers: graphical and measure-theoretic foundations support
  the model languages, which support identification, which supports the estimation and
  design methods on top. A retrieval index over all $8{,}179$ declarations spans the
  stack and is the interface the pipeline queries.}
  \label{fig:causalean}
\end{figure}

\subsection{A tour of the clusters}
\label{sec:library-tour}

The library layers from graphical and measure-theoretic foundations up to the
estimation and design methods that depend on them (\Cref{fig:causalean}).

\paragraph{Graphs and structural models.}
The Graph cluster builds directed acyclic graphs with a decidable edge
relation and a stored topological order, on top of which sit the parent, child, and
ancestor operations, $d$-separation implemented as Bayes-Ball reachability,
single-world intervention graphs, and the $c$-component decomposition that the
identification algorithm needs. The SCM cluster turns these graphs into
structural causal models. It defines interventions and the $\mathrm{do}$-operator,
proves the semi-graphoid axioms that justify do-calculus, represents factored
kernels, and develops graphical identification proper: backdoor and frontdoor
adjustment, general adjustment criteria, and a soundness proof for the ID algorithm. These results take a causal query to a formula in
the observational distribution, or to a nonidentification proof.

\paragraph{Potential outcomes and identification.}
The PO cluster holds the most definitions, because it carries the
two identification branches the pipeline uses most. Exact identification covers the
standard estimands---ATE, ATT, difference-in-differences and its Callaway--Sant'Anna
group-time refinement, LATE, regression discontinuity, proximal and dynamic-treatment
designs---each stated as an equality between a counterfactual contrast and an
estimable functional. Partial identification provides tight bounds for estimands that cannot be point-identified: Manski's worst-case bounds with their
monotone-treatment-response and instrument refinements, the sharp Balke--Pearl bounds
for a binary instrument, Lee's trimming bounds under selection, and marginal-sensitivity
models.

\paragraph{Estimation and asymptotic statistics.}
Estimation, the largest cluster by volume, is the semiparametric machinery:
double/debiased machine learning for the ATE, ATT, and CATE; efficient influence
functions obtained by projection onto a tangent space; structure-agnostic minimax
lower bounds with matching estimators; and convergence rates for nonparametric
instrumental variables. These results draw on the
Stat cluster, which formalizes the probability and empirical-process theory
underneath: central limit theorems, U-statistics with their H\'ajek projections,
Glivenko--Cantelli and bracketing-entropy tools, M- and Z-estimation, concentration
inequalities with localization, and the bootstrap. The localized theory covers star
hulls, sub-root envelopes, local Rademacher complexity, and the critical radius that
controls a localized uniform-deviation bound.

\paragraph{Panel, experiments, and discovery.}
Panel builds general machinery for panel data: potential outcomes indexed by treatment path, fixed effects, and
least squares as projection with a Frisch--Waugh--Lovell decomposition. Using that
machinery, it characterizes the estimands that two-way fixed-effects and
difference-in-differences regressions actually recover, including the
negative-weights decomposition and event-study contamination that motivate modern
DiD estimators. Experimentation covers randomization inference, Horvitz--Thompson
estimation, Neyman allocation, and central limit theorems under network interference.
Discovery formalizes identifiability results for causal structure learning, such as
LiNGAM under non-Gaussianity and invariant prediction, while ML and the local Mathlib
supplements provide the learning-theory and measure-theoretic lemmas the rest of the
library draws on.

\subsection{Flagship results}
\label{sec:library-flagship}

The library also formalizes classical results from the literature.
\Cref{tab:flagship} catalogs the main result families across its major research
areas. Each entry is a named declaration with a machine-checked proof;
\Cref{app:flagship} gives its source location.

\begin{longtable}{@{}>{\raggedright\arraybackslash}p{0.52\textwidth} >{\raggedright\arraybackslash\ttfamily\scriptsize}p{0.40\textwidth}@{}}
  \caption{Flagship theorems across \Causalean{}'s major research areas. Each entry
  is a named, machine-checked declaration; long identifiers may wrap.}
  \label{tab:flagship} \\
  \toprule
  \normalfont Result & \normalfont\rmfamily\small Declaration \\
  \midrule
  \endfirsthead
  \multicolumn{2}{@{}l}{\small\itshape Table \thetable\ continued} \\
  \toprule
  \normalfont Result & \normalfont\rmfamily\small Declaration \\
  \midrule
  \endhead
  \bottomrule
  \endfoot
  \multicolumn{2}{@{}l}{\textit{Graphical and structural causal models}} \\
  Markov equivalence iff two DAGs have the same skeleton and immoralities & markovEquiv\ub iff\ub sameSkeleton\ub sameImmoralities \\
  Rule 2 of do-calculus under its graphical condition & do\ub rule2\ub kernel \\
  Backdoor adjustment identifies an interventional distribution almost everywhere & backdoor\ub identifiable\ub ae \\
  Frontdoor adjustment identifies an interventional distribution almost everywhere & frontdoor\ub identifiable\ub ae \\
  Soundness of the graphical ID algorithm on the discrete positive class & id\ub sound\ub discrete \\
  \addlinespace
  \multicolumn{2}{@{}l}{\textit{Causal discovery}} \\
  LiNGAM identifiability from nonzero source kurtosis & lingam\ub identifiability\ub kurtosis \\
  Soundness of invariant causal prediction & icp\ub sound \\
  Completeness of invariant prediction for linear-Gaussian models & icp\ub complete\ub linearGaussian \\
  Identifiability in linear causal disentanglement & disentanglement\ub identifiability \\
  \addlinespace
  \multicolumn{2}{@{}l}{\textit{Potential outcomes, exact identification, and partial identification}} \\
  Wald-ratio identification of the local average treatment effect & late\ub wald \\
  Difference-in-differences identification of the ATT & att\ub did \\
  Callaway--Sant'Anna group-time ATT identification & att\ub csdid \\
  Sharp regression-discontinuity identification & rdd\ub identification \\
  Fuzzy regression-discontinuity identification & frd\ub identification \\
  Wald identification for dynamic treatment timing & whenToTreat\ub wald \\
  Manski worst-case ATE bounds & manski\ub bounds\ub ATE \\
  Sharpness of the Balke--Pearl bounds for a binary instrument & balkePearl\ub sharp \\
  Lee bounds for the ATT under selection & lee\ub bounds\ub ATT\ub AS \\
  Pointwise Imbens--Manski coverage for partially identified parameters & imbensManski\ub pointwise\ub coverage \\
  \addlinespace
  \multicolumn{2}{@{}l}{\textit{Estimation and statistical theory}} \\
  Asymptotic normality of the debiased-ML ATE estimator & dml\ub ATE\ub tendstoNormal \\
  Attainment of the Hahn efficiency bound by debiased-ML ATE & dml\ub ATE\ub attains\ub hahn\ub bound \\
  Asymptotic linearity of debiased-ML ATT estimation & dml\ub ATT\ub isAsymLinear \\
  Asymptotic normality of partially linear DML & plr\ub dml\ub tendstoNormal \\
  Asymptotic linearity of sequential doubly robust DTR estimation & seqDR\ub dml\ub isAsymLinear \\
  Structure-agnostic minimax lower bound for ATE estimation & minimax\ub lower\ub bound\ub var\ub causal \\
  Optimal weighting for generalized method of moments & gmm\ub efficiency \\
  Central limit theorem for regular order-$m$ U-statistics & uStatisticOrder\ub clt \\
  \addlinespace
  \multicolumn{2}{@{}l}{\textit{Panel and event-study methods}} \\
  Causal decomposition of staggered-adoption TWFE into weighted contrasts & twfe\ub po\ub decomposition \\
  Characterization of linear-unbiased BJS imputation estimators & bjs\ub linear\ub unbiased\ub iff\ub imputation\ub form \\
  Event-study pretrends induced by post-treatment effects & apparent\ub pretrends\ub from\ub post\ub treatment\ub of\ub cellGrid \\
  \addlinespace
  \multicolumn{2}{@{}l}{\textit{Experimentation under interference}} \\
  Consistency of Horvitz--Thompson estimation under unknown interference & htEst\ub consistent\ub eate \\
  Central limit theorem for a two-stage interference direct effect & directEffect\ub clt \\
  Stein-method CLT for estimators under network interference & localDependenceCLT\ub of\ub stein \\
  Wald coverage under network interference & wald\ub coverage\ub of\ub stein \\
\end{longtable}

\subsection{Retrieval}
\label{sec:library-retrieval}

A library of this size is only usable if its declarations can be found. The library provides a
keyword record for every declaration: its name, kind, module, source text,
docstring, cross-references, axiom dependencies, and whether its proof uses
\code{sorry}. This index serves readers and agents alike. The human-readable API
documentation is generated from it, and it is also what a search query is ranked
over.

For agents we build a search engine on top of it, and we train a retrieval encoder
on \Causalean{} itself to drive part of it. The training pairs come from the keyword
index: a theorem's docstring against the declarations it is built on, and each
declaration's \lean{} statement against its own docstring. The split is by module, so
the encoder is evaluated only on modules it never saw. All $8{,}179$ declarations
are then embedded into a vector index of $1{,}024$ dimensions.

Search runs in three modes. A concept mode expands a natural-language query with
causal-inference synonyms. A type-pattern mode matches a fragment of \lean{} syntax
against the symbols in each declaration's statement. A goal-directed mode ranks
candidates against the goal a proof is stuck on. Concept and goal queries are ranked
twice, once by keyword score and once by embedding similarity, and the two rankings
are combined by reciprocal-rank fusion. A fine-tuned cross-encoder then reorders the
leading concept candidates. Type-pattern
queries use the keyword index alone.

\subsection{Axiom checks}
\label{sec:library-axiom-checks}

The whole library is \code{sorry}-free, and none of the $8{,}179$ declarations
rests on a hand-written \code{axiom}. The only non-standard axioms come from
\code{native\_decide}, a tactic that settles a finite computation by running the
compiled code and trusting the result; these occur in finite-graph decidability
arguments and several minimax calculations, and we report them explicitly.

Two parts of the library are not original to \Causalean{}. Both are adapted copies
of external \lean{} projects, bumped to \Causalean{}'s \lean{}/Mathlib pin. The
first is the Karush--Kuhn--Tucker first-order necessary conditions under LICQ and
affine constraint qualifications, taken from OptSuite's
\code{optlib}~\citep{optlib2025} and consumed by the Estimation cluster's minimax
lower bounds. The second collects Rademacher complexity, McDiarmid's inequality,
symmetrization, and the Dudley entropy integral, taken from
\code{lean-rademacher}~\citep{leanrademacher2025} and consumed across Stat,
Estimation, and ML.

\section{The \CausalSmith{} Pipeline}
\label{sec:pipeline}

\CausalSmith{} is a research pipeline with four stages:
Discovery, Formalization, Proof Construction, and Presentation. The pipeline either
accepts a researcher-supplied topic or selects one itself, then proposes a causal-inference
result in natural language, formalizes it, and produces a paper
linked to the \lean{} statements. New results are stored as dependency graphs throughout the pipeline. Discovery creates this graph; formalization
maps its nodes to planned \lean{} declarations; proof construction fills those
declarations and updates their dependencies; statement matching checks each formal
node against the intended claim; and presentation writes the paper from the
reviewed graph.

The pipeline runs the stages in a fixed order and retries a failed stage within
fixed limits. An orchestrator launches it and records its verdicts. \Cref{fig:pipeline} summarizes the workflow;
\Cref{app:pipeline-spec,app:pipeline-proof-loop,app:pipeline-paper,app:study}
give implementation details of the stages, the proof and audit loop, the
presentation workflow, and study mode.

\subsection{The Dependency Graph as a Persistent Run Record}
\label{sec:pipeline-graph}

Lean blueprints and autoformalization pipelines already represent a development as a
dependency graph of statements, each node carrying a formalization
status~\citep{leanarchitect2026,proofflow2025}. We adopt that structure and extend the
node. Each result is stored as an acyclic directed graph whose nodes are
statements---a setup, a definition, an assumption, a lemma, or the headline
theorem---and whose edges represent dependencies. Alongside its \lean{} declaration, a
node carries the intended natural-language statement and a review status recording
whether the two have been compared and whether they matched, together with a class
that separates two kinds of dependency. A must-prove node is a result the paper
proves, and a cited node is a result taken from the literature, carried with its
source and lying outside the paper's novelty. The graph is therefore both a plan and
an audit record. A typical completed result has on
the order of a hundred nodes and a few hundred edges (\Cref{fig:graph});
\Cref{app:graph-schema} gives the full schema.

Within Discovery, the agents write the new results as natural-language nodes with their
dependency edges. Formalization then attaches the intended
\lean{} declarations to each node. Proof construction adds the proof dependencies and fills in the \lean{} proof. Statement matching compares the natural-language statement with the \lean{} declaration for each node and updates the review status. Presentation
then uses the reviewed graph to link each part of the paper to its code. When a statement changes, the pipeline marks the downstream nodes unreviewed again and rechecks them.

\subsection{Discovery, formalization, and proof}
\label{sec:pipeline-discovery-proof}

The pipeline supports two entry modes. A researcher may supply a topic anchor, or the
main orchestrator may invoke the topic-selection skill as the first part of
Discovery. The selector searches recent work and previous runs, and reads theoretical papers together with their citing and follow-up literature in full, to find inspiration and propose a few candidate topics. It then ranks them by mathematical potential and by whether the result would have a concrete application, and submits the top candidate to an independent adversarial topic review. The pipeline is started only when the review passes.

Once a topic is fixed, the remaining discovery stages derive the result in natural
language and build the graph. A proposer drafts a question and an informal solution; a
novelty-and-duplication review compares it with the literature and judges whether it is
novel. Then a solver solves the proposed open questions in natural language. Their
outputs become the initial graph nodes and edges. When a proposed claim is too
strong or too weak, the solver may adjust it to obtain a meaningful result. A mathematically
sound but insufficiently novel proposal is downgraded, whereas an incorrect or trivial proposal is rejected.

Formalization turns this graph into a proof plan. Each node receives an
intended \lean{} declaration, a module location, and either a statement to prove or
an existing \Causalean{} result to reuse. The plan becomes a skeleton file in which
each statement to prove is left \code{sorry}. A review-and-fill loop then closes those open statements. On each iteration, the reviewer first checks that every declaration remains aligned with the graph and repairs declarations whose statements are not faithful; it then fills each
remaining \code{sorry}, using the search engine of \Cref{sec:library-retrieval} to reuse existing results. An automatic check flags hypotheses the proof never uses, which may indicate a vacuous or mis-stated theorem.

\subsection{Statement matching}
\label{sec:pipeline-audit}
The statement-match review in \Cref{fig:pipeline} is where the pipeline guarantees
faithfulness. As \Cref{sec:prelim} notes, the \lean{} type check says nothing about
whether the formal statement is the one the researcher meant to prove. A formalization can carry no proof gaps and still fail expert
review, because a definition is too narrow, a hypothesis is vacuous, or an unproven
step has been promoted to an axiom. This review
checks each formal statement against the intended claim stored in its graph node. The
main criterion is logical equivalence: the two must have equivalent assumptions and
equivalent conclusions, with neither stronger nor weaker than the other.
\Cref{tab:taxonomy} groups the failure modes into four families and records where the
pipeline catches each one.

\begin{table}[t]
  \centering
  \small
  \caption{Failures that can pass the \lean{} type check while changing the intended claim.}
  \label{tab:taxonomy}
  \begin{tabular}{@{}p{0.15\textwidth}p{0.46\textwidth}p{0.31\textwidth}@{}}
    \toprule
    Family & Failure mode & How the review catches it \\
    \midrule
    Wrong statement & Type-checks but states the wrong proposition. & Per-node comparison with the intended claim. \\
    Vacuity & Concept defined as \code{True} and a witness that carries no real obstruction. & A witness must carry a real obstruction and cannot be \code{True} or an empty structure; satisfiability is screened earlier, by the mathematical review of the proposal. \\
    Unproved step & Step asserted as \code{axiom}; \code{sorry}/\code{admit}; a node the proof depends on mis-tagged as cited. & Automatic scan for these keywords; source check for cited nodes. \\
    Over-narrow statement & Dropped hypothesis; hardcoded constant where general was intended. & Comparison with the intended claim; check for unused hypotheses. \\
    \bottomrule
  \end{tabular}
\end{table}

The review runs in two layers. The first is mechanical. A scan rejects any completed
artifact that contains \code{axiom}, \code{sorry}, \code{admit}, \code{opaque}, or \code{unsafe}, and it applies the same test
to every library module the artifact depends on, directly or indirectly. The build itself
rejects any artifact with an unfinished proof.
In the second layer, several agents read the \lean{} declaration text extracted
from the compiled file, compare it against the intended statement, and flag
any divergence as drift; the review requires every claimed witness to
carry a concrete obstruction and every cited node to have a real source
match.

Once every statement has a proof, two independent models recheck the full frozen
graph, including nodes the latest changes did not touch, comparing every statement
against the intended claim without reference to the earlier review. This is the final review in \Cref{fig:pipeline} before the result enters the presentation stage.

\subsection{Library feedback}
\label{sec:pipeline-study}

Two feedback channels connect one run to future runs: a run record of what has been attempted, and a verified library of what has been proved.

The run record preserves every run regardless of its outcome, from accepted to
rejected, together with its proposal, mathematical core, and review verdicts
(\Cref{tab:funnel,app:repro}). Before a new proposal is drafted, Discovery
searches these records for open gaps and near-duplicates, and the
novelty-and-duplication review compares a fresh proposal against prior entries. Failed
and downgraded runs contribute alongside accepted ones: a documented rejection
prevents a later run from repeating the same dead end, and a downgraded result
sharpens the novelty target for a related question.

The library channel handles reusable lemmas or theorems during the run. When the graph identifies a missing reusable lemma, the pipeline can prove that lemma and promote it to
\Causalean{}. A library builder (called \code{study} mode in our pipeline) first plans the lemmas and writes a file with the statements left open. A prover then fills in the proofs against the live compiler, and a reviewer assesses whether the resulting lemma is generic, reusable, non-vacuous, and \code{sorry}-free. A build check confirms this assessment. \Cref{app:study} describes study mode in full.

\subsection{Presentation}
\label{sec:pipeline-presentation}

The presentation stage converts an accepted result
into a working paper in which each theorem, definition, and assumption links to its
verified \lean{} source. Linking prose to code is established
practice~\citep{leanarchitect2026}. The link records both facts attached to a graph node: the \lean{} declaration compiles, and the statement has been checked as matching the English claim. An equivalence check applies the criterion of \Cref{sec:pipeline-audit}: the paper statement and the \lean{} declaration must express the same proposition, with neither stronger nor weaker than the other.

An LLM reviewer at the end of this stage reads the assembled paper the way a
journal referee would, treating the verified theorems as given, and judges the
significance of the contribution, whether each prose claim matches the verified
statement in strength and scope, and the exposition, clarity, and positioning against
related work. It returns a recommendation with a score, and findings on what could be improved. The score is advisory and is recorded in the run record.

The same stage publishes the result to the companion site. There the paper carries a
proof map in the margin, showing which of the paper's own results each proof invokes.
Every statement links to the \lean{} declaration it was audited against, hypothesis by
hypothesis. A reader signed in with a GitHub account can verify a single statement, rate
the paper or one of its results, and comment on any passage. The stage also builds a
short slide deck from the same graph. \Cref{app:site} shows each of these.

\section{Results}
\label{sec:eval}

We analyze the pipeline's run records and logs in this section. We give the distribution of outcomes across runs, report which kinds of
self-proposed question the system converts into accepted results, examine one
accepted result in depth, report the axiom audit that screens each accepted
result, and trace the library feedback loop closing on a concrete lemma.

All results reported here come from one base-model lineup, whose members were
updated during the campaign. OpenAI reasoning models carry the main mathematics and
the \lean{} formalization: GPT-5.5 until 2026-07-10 and GPT-5.6 Sol thereafter.
Anthropic's Claude handles \lean{} code review and planning throughout, as Opus~4.7
until 2026-05-28, Opus~4.8 until 2026-07-24, and Opus~5 after. Nine of the fourteen
accepted runs ran entirely under Opus~4.8 and the other five entirely under Opus~5. Mechanical stages
run on GPT-5.6 Terra, and the presentation write-ups are drafted by GPT-5.5.
\Cref{app:repro} maps every role to its model and records what the run logs attest.

The run record holds $144$ runs, distributed across three outcomes
(\Cref{tab:funnel}). A run is accepted only when it is sound, novel at its
requested publishable tier, and proved to completion in \lean{}; fourteen runs meet this bar. The remaining runs are retained: as \Cref{sec:pipeline-study} describes, downgraded
and failed runs feed the record that screens later proposals.

\begin{table}[t]
  \centering
  \small
  \caption{The run catalog: $144$ recorded \CausalSmith{} runs by outcome. A
  separate literature-reproduction track adds one reproduced partial-identification
  result. Counts are entries in the run record; a re-run that supersedes an earlier
  attempt at the same question counts separately, and nine legacy runs from the
  retired proposal track that predates the current pipeline are excluded.}
  \label{tab:funnel}
  \begin{tabular}{@{}lrl@{}}
    \toprule
    Outcome & Runs & Review that determined it \\
    \midrule
    Accepted   & $14$  & sound, novel at tier, proved in \lean{} \\
    Downgraded & $59$  & sound, below the novelty target \\
    Failed     & $71$  & rejected at the proposal or the mathematics \\
    \midrule
    Total      & $144$ & \\
    \bottomrule
  \end{tabular}
\end{table}

\subsection{Outcomes by cluster and question type}
\label{sec:eval-question-type}

Discovery files every run under one of six clusters before any mathematics begins.
Sorted that way (\Cref{tab:by-cluster}), the accepted results gather in \code{Stat},
\code{Experimentation}, and \code{Panel}. By contrast, the three identification and
structural-model clusters together hold $94$ runs and only one acceptance.

\begin{table}[!htb]
  \centering
  \small
  \caption{The run catalog by the cluster Discovery assigned at proposal time.
  ``Accept rate'' is accepted runs over total runs in the cluster.}
  \label{tab:by-cluster}
  \begin{tabular}{@{}lrrrrr@{}}
    \toprule
    Cluster & Acc. & Down. & Fail. & Total & Accept rate \\
    \midrule
    \code{Stat}            & $8$ & $18$ & $5$  & $31$ & $26\%$ \\
    \code{Experimentation} & $4$ & $7$  & $1$  & $12$ & $33\%$ \\
    \code{Panel}           & $1$ & $0$  & $6$  & $7$  & $14\%$ \\
    \midrule
    \code{ExactID}         & $1$ & $8$  & $27$ & $36$ & $3\%$ \\
    \code{PartialID}       & $0$ & $24$ & $30$ & $54$ & $0\%$ \\
    \code{SCM}             & $0$ & $2$  & $2$  & $4$  & $0\%$ \\
    \midrule
    Total                  & $14$ & $59$ & $71$ & $144$ & $10\%$ \\
    \bottomrule
  \end{tabular}
\end{table}

We also labeled each run by what its proposal is about, using the following rules. The catalog is shown in \Cref{tab:by-type}.

\begin{itemize}[leftmargin=1.4em,itemsep=1pt,topsep=2pt]
  \item \textbf{Method comparison.} The topic compares two or more published estimators or
  identification frameworks and asks how they differ: non-equivalence,
  strict-extension, or equivalence.
  \item \textbf{Assumption relaxation.} The topic weakens a named assumption, under a
  budget on how far it may be violated, and asks for the identification result---an
  identified set, a threshold, a phase transition---as a function of that budget.
  Sensitivity and ambiguity models are the common instances.
  \item \textbf{Gap filling.} The setting already exists in the literature,
  but some gaps are left open: a matching bound; an exact rate, order, or constant; a minimax optimum or optimal design; or an explicit identifying formula.
  \item \textbf{New setting.} Anything else: a newly posed object, criterion, algorithm, or framework, with no published counterpart to be matched against.
\end{itemize}

\begin{table}[!htb]
  \centering
  \small
  \caption{The run catalog by question type. Types are assigned by the rule in the
  text, applied to the proposal-time topic string each recorded run carries. We assigned the labels once the runs had finished (\Cref{app:repro}).}
  \label{tab:by-type}
  \begin{tabular}{@{}lrrrrr@{}}
    \toprule
    Question type & Acc. & Down. & Fail. & Total & Accept rate \\
    \midrule
    Gap filling           & $14$ & $22$ & $8$  & $44$  & $32\%$ \\
    \midrule
    Assumption relaxation & $0$  & $29$ & $22$ & $51$  & $0\%$ \\
    New setting           & $0$  & $6$  & $28$ & $34$  & $0\%$ \\
    Method comparison     & $0$  & $2$  & $13$ & $15$  & $0\%$ \\
    \midrule
    Total                 & $14$ & $59$ & $71$ & $144$ & $10\%$ \\
    \bottomrule
  \end{tabular}
\end{table}

All fourteen acceptances are gap filling. The remaining
three types each fail in a characteristic way. Assumption-relaxation runs
usually reach a correct derivation and then lose on novelty, giving $29$
downgrades against $22$ failures; reviewers describe what comes back as
``generic Manski-style outer containment'' or ``while correct, a standard
finite-dimensional support/projection calculation,'' so the difficulty lies in
conception rather than in the proof.
Method-comparison and new-setting runs stop earlier, $13$ of $15$ and $28$ of
$34$ failing outright, often at the proposal stage. Gap-filling proposals often have a more quantitative target, such as a better convergence rate or a matched lower bound, while the other
three types do not.

\paragraph{Where the gaps come from.}
Some gaps are explicitly stated in the literature, and some the system finds for itself. We read the anchor
paper behind each of the fourteen accepted results. Three inherit a gap the anchor
states in its own text: \citet{zeng2024discrete} leave both the closure of their
$\log^2$ gap and the sharp minimax dependence on their homogeneity radius to future
work, and \citet{cortez2023snipe} trace the degree gap between their bounds to ``a
weakness in the analyses.'' The other eleven aim at a quantity the literature had yet
to ask about, such as the pseudo-true estimand of staggered
PPML~\citep{moreaukastler2025ppml}, a differentially private counterpart of a known
conditional average treatment effect rate~\citep{kennedy2024minimaxcate}, or the
minimax confidence-set length for a transported complier average causal
effect~\citep{chen2025generalizing}.

\subsection{A flagship result: closing a minimax gap for the ATE}
\label{sec:eval-flagship}

The strongest accepted result closes a standing gap in the minimax theory of
average-treatment-effect (ATE) estimation under high-dimensional discrete
confounding. \citet{zeng2024discrete} established a minimax lower bound of order
$n^{-1}+\big(d/(n\log n)\big)^2$ for estimating the ATE from $n$ i.i.d.\ observations
$(X,A,Y)$ with a discrete confounder $X\in\{1,\dots,d\}$, binary treatment $A$ and
outcome $Y$, and strict-interior overlap $\epsilon\le\Pr(A=1\mid X=k)\le 1-\epsilon$.
The estimators they analyze, however, leave a $\log^2$-factor gap between the known
upper and lower bounds. The run constructs an estimator that closes
it.

Under the stated overlap, consistency, and conditional-exchangeability assumptions,
the target is the adjustment functional
$\tau(P)=\sum_{k=1}^{d}p_k(\mu_{1k}-\mu_{0k})=\mathbb{E}[Y(1)-Y(0)]$, where
$p_k=\Pr(X=k)$ and $\mu_{ak}=\mathbb{E}[Y\mid A=a,X=k]$. The estimator splits the
sample in two and uses a pilot count to label each covariate cell heavy or light. The
two regimes are then handled differently. Heavy, well-populated cells receive a
plug-in ratio estimator, and light, sparsely sampled cells receive a best-polynomial
approximation of the cell functional with unbiased factorial-moment lifting. The headline theorem states that, for every fixed overlap $0<\epsilon<1/2$, this estimator attains
\[
  \mathsf R_{n,d,\epsilon}\;\asymp_\epsilon\;
  \frac1n+\Big(\frac{d}{n\log n}\Big)^2
  \qquad\text{uniformly for } d\lesssim_\epsilon n\log n,
\]
so the minimax MSE has parametric order $n^{-1}$ whenever $d=O(\sqrt n\log n)$ and
tends to zero if and only if $d=o(n\log n)$.

The novel part is the hybrid estimator and its matched rate. The minimax lower half is not checked here; it is cited from the published moment-matching bound of
\citet{zeng2024discrete}. In the \lean{} code, this imported bound appears as
an explicit hypothesis on the headline theorem, which follows the
cited-result discipline of \Cref{tab:taxonomy}: an external result the proof depends on is
carried as a hypothesis with source evidence, so a reader sees which part is
proved in this work and which is from the literature.

\subsection{Machine-checked soundness}
\label{sec:eval-axiom-checks}

Each accepted result is screened for the unproved-step failures of
\Cref{tab:taxonomy} before it is recorded. Starting from a clean build, we ask \lean{}
which axioms the headline theorem depends on, and we cross-check the answer by
scanning the recorded module and every library module it reaches for \code{sorry},
\code{admit}, and \code{axiom} (\Cref{app:repro}). For the all accepted papers,
the formalization contain none of these, and the theorems reduces to \lean{}'s standard axioms; its single external input, the
\citet{zeng2024discrete} lower bound, is visible as a hypothesis.

\subsection{Library feedback in action}
\label{sec:eval-selfimprove}

The flagship result of \Cref{sec:eval-flagship} gives a complete instance of the
library feedback loop. Its estimator labels each covariate cell heavy or light
by a pilot count, so the proof needs two-sided multiplicative Chernoff tails for
the count of an i.i.d.\ $[0,1]$-valued statistic, which Mathlib does not have. The run proved them, and promotion moved them into \Causalean{} as
\code{Causalean.Stat.Concentration.}\allowbreak\code{TailBounds.BinomialCount},
whose \code{bernoulliCount\_upper\_tail} and \code{bernoulliCount\_lower\_tail}
the run then applies.

The next accepted result on the same estimand consumed that promotion. The two-sided
minimax bracket for the discrete ATE under a shrinking heterogeneity radius needs the
pilot argument at greater generality, over an arbitrary finite category set where the
flagship had two cell types. The run then built
\code{Causalean.Stat.SampleSplit.}\allowbreak\code{FiniteCategoryPilot}, which imports
\code{BinomialCount} and reuses its moment-generating-function bound. So a run required a lemma, the
system proved it, promotion added it to the library, and a later run on the same
problem built on it instead of repeating it. Twenty-eight further library builds have
produced promotions of their own, including Fano's inequality, a Bretagnolle--Huber affinity bound for
arbitrarily many hypotheses that a dose-response converse needed and a later Le Cam
two-point reduction reuses, a KL density-tilt expansion, and the Ehlich--Zeller and
Bernstein--Szeg\H{o} tools behind an accepted Chebyshev rollout design. Estimating how
library growth trades off against cost per result remains future work
(\Cref{sec:discussion}).

\section{Discussion and Limitations}
\label{sec:discussion}

Automated theoretical research needs a reliable basis for judging correctness, and LLM
reviewers can be misled. \CausalSmith{} answers that at three separate levels of
confidence. The \lean{} type check settles whether a proof is valid, and the statement
audit guarantees that the \lean{} statement matches the intended claim. Whether a finished paper matters stays a human judgment. Keeping the three apart is what stops a valid proof from being read as an important result.

Most of the pipeline is domain-general, and only the formal library is specific to
causal inference. Moving to another field means building or adopting a comparable
library and retrieval interface. The graph, the audit, and the library feedback loop
should carry over with little change.

Our current evaluation has the following limitations. First, research quality is rated by the pipeline itself, and these LLM ratings do not have independent human validation. Second, the statement-match audit is itself an LLM task; frontier models handled it well in our observations, but we do not estimate how often they err.

Taken together, the verified library, the self-directed pipeline, and the feedback
loop that grows the library give machine-checked proofs of self-proposed results with
audited statements. We leave better novelty-evaluation metrics and faithfulness-audit benchmarks to future work.

\bibliographystyle{plainnat}
\bibliography{references}

\appendix
\section{Flagship theorem locations}
\label{app:flagship}

\Cref{tab:flagship-loc} lists the \lean{} declaration name and source location of
every flagship result named in \Cref{sec:library-flagship}, so that a reader can look
each one up directly. Paths are relative to the \code{Causalean/} package root, at the
pinned toolchain \code{leanprover/lean4:v4.29.0-rc3}.

\begin{longtable}{@{}>{\raggedright\arraybackslash\ttfamily\scriptsize}p{0.41\textwidth} >{\raggedright\arraybackslash\ttfamily\scriptsize}p{0.48\textwidth}r@{}}
  \caption{Source locations of the flagship declarations in \Cref{tab:flagship}.}
  \label{tab:flagship-loc} \\
  \toprule
  \normalfont\small Declaration & \normalfont\small File & \normalfont\small Line \\
  \midrule
  \endfirsthead
  \multicolumn{3}{@{}l}{\small\itshape Table \thetable\ continued} \\
  \toprule
  \normalfont\small Declaration & \normalfont\small File & \normalfont\small Line \\
  \midrule
  \endhead
  \bottomrule
  \endfoot
  markovEquiv\_iff\_sameSkeleton\_sameImmoralities & Graph/MarkovEquiv.lean & 51 \\
  do\_rule2\_kernel & SCM/Do/DoCalculus.lean & 97 \\
  backdoor\_identifiable\_ae & SCM/ID/Backdoor.lean & 377 \\
  frontdoor\_identifiable\_ae & SCM/ID/Frontdoor.lean & 1190 \\
  id\_sound\_discrete & SCM/ID/GraphicalThms/IDSoundDiscrete.lean & 30 \\
  lingam\_identifiability\_kurtosis & Discovery/LiNGAM/LiNGAMKurtosis.lean & 69 \\
  icp\_sound & Discovery/InvariantPrediction/Soundness.lean & 34 \\
  icp\_complete\_linearGaussian & Discovery/InvariantPrediction/\ldots/Completeness.lean & 270 \\
  disentanglement\_identifiability & Discovery/LinearDisentanglement/Identifiability.lean & 35 \\
  late\_wald & PO/ID/Exact/LATE.lean & 413 \\
  att\_did & PO/ID/Exact/DID.lean & 150 \\
  att\_csdid & PO/ID/Exact/CSDID.lean & 423 \\
  rdd\_identification & PO/ID/Exact/RDD/SharpRDD.lean & 295 \\
  frd\_identification & PO/ID/Exact/RDD/FuzzyRDD.lean & 575 \\
  whenToTreat\_wald & PO/ID/Exact/DynamicLATE/WhenToTreat.lean & 332 \\
  manski\_bounds\_ATE & PO/ID/Partial/Manski/NonAsp.lean & 72 \\
  balkePearl\_sharp & PO/ID/Partial/BalkePearl/Sharp.lean & 852 \\
  lee\_bounds\_ATT\_AS & PO/ID/Partial/Lee/Main.lean & 33 \\
  imbensManski\_pointwise\_coverage & PO/ID/Partial/Inference/ImbensManski.lean & 171 \\
  dml\_ATE\_tendstoNormal & Estimation/ATE/DML.lean & 899 \\
  dml\_ATE\_attains\_hahn\_bound & Estimation/Efficiency/ATEVariance.lean & 415 \\
  dml\_ATT\_isAsymLinear & Estimation/ATT/DML.lean & 237 \\
  plr\_dml\_tendstoNormal & Estimation/PLR/DML.lean & 158 \\
  seqDR\_dml\_isAsymLinear & Estimation/DTR/DTRInstance.lean & 145 \\
  minimax\_lower\_bound\_var\_causal & Estimation/MinimaxATE/Causal/Minimax.lean & 172 \\
  gmm\_efficiency & Stat/GMM/VarianceAlgebra.lean & 106 \\
  uStatisticOrder\_clt & Stat/UStatistic/OrderM/CLT.lean & 52 \\
  twfe\_po\_decomposition & Panel/\ldots/CausalDecomposition.lean & 81 \\
  bjs\_linear\_unbiased\_iff\_imputation\_form & Panel/\ldots/ImputationEventStudy/PanelBridge.lean & 370 \\
  apparent\_\allowbreak pretrends\_\allowbreak from\_\allowbreak post\_\allowbreak treatment\_\allowbreak of\_\allowbreak cellGrid & Panel/\ldots/EventStudyContamination/Contamination.lean & 63 \\
  htEst\_consistent\_eate & Experimentation/UnknownInterference/Consistency.lean & 133 \\
  directEffect\_clt & Experimentation/\ldots/Asymptotic/CLT.lean & 118 \\
  localDependenceCLT\_of\_stein & Experimentation/\ldots/SteinInstance.lean & 176 \\
  wald\_coverage\_of\_stein & Experimentation/\ldots/SteinInstance.lean & 252 \\
\end{longtable}

\section{Logic-graph schema}
\label{app:graph-schema}

Each result carries one dependency graph (\Cref{sec:pipeline-graph}), stored on disk as a
\code{FormalizationGraph}. A node has a kind, one of \code{setup},
\code{definition}, \code{assumption}, \code{lemma}, \code{theorem}, or \code{gate}. It
also holds its natural-language statement, its \lean{} declaration, a review status
(\code{unreviewed}, \code{matched}, or \code{drift}), and a
\code{gate\ub class} field, either \statusgated{} or \statuscited{}, the must-prove and
cited classes of \Cref{sec:pipeline-graph}. An assumption node takes one further label, saying
whether the assumption refines the claim, does regularity bookkeeping, or stands in for
a library gap. Edges are typed as \code{statement-uses}, \code{proof-uses}, or
\code{setup-of}, and store their two endpoints. Before a graph is written, a validator
checks it: duplicate nodes and edges are rejected, and any structural invariant that
has been violated is reported.

\section{Implementation details of the pipeline}
\label{app:pipeline-spec}

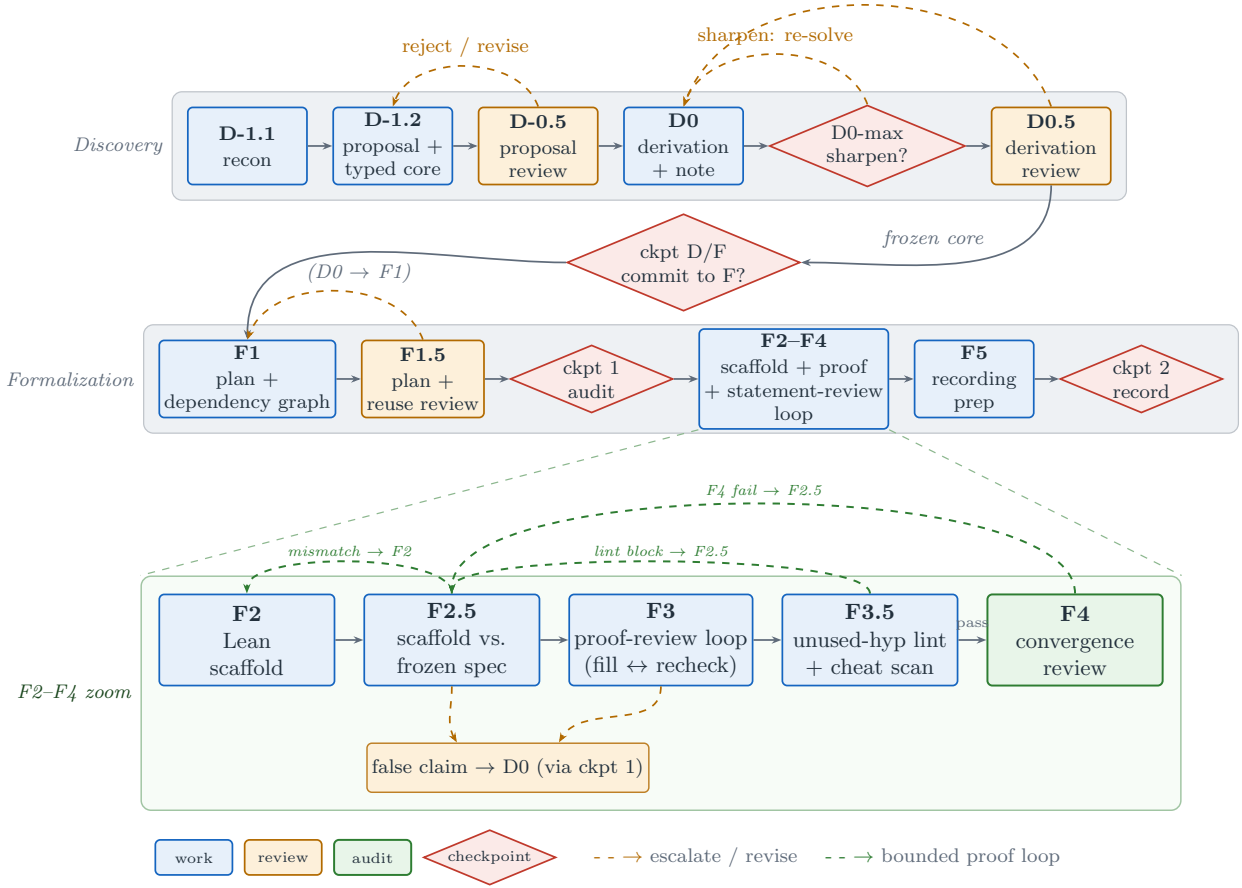
\begin{figure*}[!htb]
  \centering
  \resizebox{\textwidth}{!}{%
  \begin{tikzpicture}[
      font=\scriptsize, node distance=3.5mm and 3.5mm,
      work/.style={rounded corners=2pt, draw=cBlue, line width=0.7pt, fill=cBlueBg,
                    minimum height=11mm, minimum width=17mm, align=center, text=cInk,
                    inner sep=2pt},
      gate/.style={rounded corners=2pt, draw=cAmber, line width=0.7pt, fill=cAmberBg,
                    minimum height=11mm, minimum width=17mm, align=center, text=cInk,
                    inner sep=2pt},
      ckpt/.style={diamond, aspect=2.4, draw=cRed, line width=0.7pt, fill=cRedBg,
                    minimum height=8mm, align=center, text=cInk,
                    inner sep=1pt, font=\scriptsize},
      band/.style={rounded corners=4pt, draw=cGray!40, line width=0.5pt,
                    fill=cGrayBg, inner sep=6pt},
      sub/.style={rounded corners=2pt, draw=cBlue, line width=0.7pt, fill=cBlueBg,
                   minimum height=13mm, minimum width=25mm, align=center, text=cInk,
                   inner sep=2pt, font=\footnotesize},
      audit/.style={rounded corners=2pt, draw=cGreen, line width=0.8pt, fill=cGreenBg,
                     minimum height=13mm, minimum width=25mm, align=center, text=cInk,
                     inner sep=2pt, font=\footnotesize},
      side/.style={rounded corners=2pt, draw=cAmber!85, line width=0.6pt, fill=cAmberBg,
                    minimum height=7mm, minimum width=28mm, align=center, text=cInk,
                    font=\scriptsize, inner sep=2pt},
      zoomband/.style={rounded corners=4pt, draw=cGreen!45, line width=0.6pt,
                        fill=cGreenBg!35, inner sep=7pt},
      fl/.style={-{Stealth[length=4pt]}, line width=0.7pt, draw=cGray},
      back/.style={-{Stealth[length=4pt]}, line width=0.7pt, draw=cAmber, dashed},
      loop/.style={-{Stealth[length=4pt]}, line width=0.8pt, draw=cGreen, dashed},
      lens/.style={draw=cGreen!55, line width=0.5pt, dashed},
    ]

    \node[work] (d11) {\textbf{D-1.1}\\[1pt]recon};
    \node[work, right=of d11] (d12) {\textbf{D-1.2}\\[1pt]proposal +\\typed core};
    \node[gate, right=of d12] (dm05) {\textbf{D-0.5}\\[1pt]proposal\\review};
    \node[work, right=of dm05] (d0) {\textbf{D0}\\[1pt]derivation\\+ note};
    \node[ckpt, right=of d0] (c0) {D0-max\\{\scriptsize sharpen?}};
    \node[gate, right=of c0] (d05) {\textbf{D0.5}\\[1pt]derivation\\review};

    \node[work, below=22mm of d11] (f1) {\textbf{F1}\\[1pt]plan +\\dependency graph};
    \node[gate, right=of f1] (f15) {\textbf{F1.5}\\[1pt]plan +\\reuse review};
    \node[ckpt, right=of f15] (c1) {ckpt~1\\{\scriptsize audit}};
    \node[work, right=of c1, minimum width=23mm] (f234)
      {\textbf{F2--F4}\\[1pt]scaffold + proof\\ + statement-review\\ loop};
    \node[work, right=of f234] (f5) {\textbf{F5}\\[1pt]recording\\prep};
    \node[ckpt, right=of f5] (c2) {ckpt~2\\{\scriptsize record}};

    \begin{scope}[on background layer]
      \node[band, fit=(d11)(d05), label={[font=\scriptsize\itshape, text=cGray]left:\;Discovery}] {};
      \node[band, fit=(f1)(c2),   label={[font=\scriptsize\itshape, text=cGray]left:\;Formalization}] (fband) {};
    \end{scope}

    \draw[fl] (d11)  -- (d12);
    \draw[fl] (d12)  -- (dm05);
    \draw[fl] (dm05) -- (d0);
    \draw[fl] (d0)   -- (c0);
    \draw[fl] (c0)   -- (d05);

    \node[ckpt] (cDF) at ($(d0.south) + (0,-11mm)$)
      {ckpt~D/F\\{\scriptsize commit to F?}};

    \draw[fl] (d05.south) to[out=-90,in=0]
      node[above=1mm, font=\scriptsize\itshape, text=cGray, pos=0.6] {frozen core}
      (cDF.east);
    \draw[fl] (cDF.west) to[out=180,in=90]
      node[below, font=\scriptsize\itshape, text=cGray, pos=0.5] {(D0 $\to$ F1)}
      (f1.north);

    \draw[fl] (f1)   -- (f15);
    \draw[fl] (f15)  -- (c1);
    \draw[fl] (c1)   -- (f234);
    \draw[fl] (f234) -- (f5);
    \draw[fl] (f5)   -- (c2);

    \draw[back] (dm05.north) to[out=110,in=70]
      node[above, font=\scriptsize, text=cAmber, pos=0.5] {reject / revise}
      (d12.north);
    \draw[back] (c0.north)   to[out=110,in=70]
      node[above, font=\scriptsize, text=cAmber, pos=0.5] {sharpen: re-solve}
      (d0.north);
    \draw[back] (d05.north)  to[out=110,in=70] (d0.north);
    \draw[back] (f15.north)  to[out=110,in=70] (f1.north);


    \node[sub, anchor=north west] (f2z) at ($(f1.south west)+(0,-25mm)$)
      {\textbf{F2}\\ Lean\\ scaffold};
    \node[sub, right=4mm of f2z] (f25)
      {\textbf{F2.5}\\ scaffold vs.\\ frozen spec};
    \node[sub, right=4mm of f25] (f3)
      {\textbf{F3}\\ proof-review loop\\ (fill $\leftrightarrow$ recheck)};
    \node[sub, right=4mm of f3] (f35)
      {\textbf{F3.5}\\ unused-hyp lint\\ + cheat scan};
    \node[audit, right=4mm of f35] (f4)
      {\textbf{F4}\\ convergence\\ review};

    \node[side, below=8mm of f25, xshift=8mm] (escD0)
      {false claim $\to$ D0 (via ckpt~1)};

    \begin{scope}[on background layer]
      \node[zoomband, fit=(f2z)(f4)(escD0),
            label={[font=\scriptsize\itshape, text=cGreen!55!black]left:%
                   \;F2--F4 zoom}] (zpanel) {};
    \end{scope}

    \draw[fl] (f2z) -- (f25);
    \draw[fl] (f25) -- (f3);
    \draw[fl] (f3)  -- (f35);
    \draw[fl] (f35) -- node[above, font=\tiny, text=cGray] {pass} (f4);

    \draw[loop] (f25.north) .. controls +(0,6mm) and +(0,6mm) ..
      node[above=-0.5mm, font=\tiny\itshape, text=cGreen, pos=0.5]
        {mismatch $\to$ F2}
      (f2z.north);
    \draw[loop] (f35.north) .. controls +(0,6mm) and +(0,6mm) ..
      node[above=-0.5mm, font=\tiny\itshape, text=cGreen, pos=0.5]
        {lint block $\to$ F2.5}
      (f25.north);
    \draw[loop] (f4.north) .. controls +(0,17mm) and +(0,17mm) ..
      node[above=-0.5mm, font=\tiny\itshape, text=cGreen, pos=0.5]
        {F4 fail $\to$ F2.5}
      (f25.north);

    \draw[back] (f25.south) to[out=-90,in=120]
      ($(escD0.north west)!0.32!(escD0.north east)$);
    \draw[back] (f3.south) to[out=-90,in=60]
      ($(escD0.north west)!0.68!(escD0.north east)$);

    \draw[lens] (f234.south west) -- (zpanel.north west);
    \draw[lens] (f234.south east) -- (zpanel.north east);

    \node[work, anchor=north west, minimum height=5mm, minimum width=11mm,
          font=\tiny, inner sep=1pt]
          (lw) at ($(zpanel.south west)+(2mm,-4mm)$) {work};
    \node[gate, right=1.5mm of lw, minimum height=5mm, minimum width=11mm,
          font=\tiny, inner sep=1pt] (lg) {review};
    \node[audit, right=1.5mm of lg, minimum height=5mm, minimum width=11mm,
          font=\tiny, inner sep=1pt] (la) {audit};
    \node[ckpt, right=1.5mm of la, minimum height=5mm, minimum width=13mm,
          font=\tiny] (lc) {checkpoint};
    \node[right=4mm of lc, font=\scriptsize, text=cGray, align=left]
      {\textcolor{cAmber}{- - $\to$} escalate / revise \quad
       \textcolor{cGreen}{- - $\to$} bounded proof loop};
  \end{tikzpicture}%
  }
  \caption{Stage flow of a \CausalSmith{} theorem run, with an inset zoom on the
  F2--F4 proof and statement-review loop. Top band: Discovery produces a proposal and
  a mathematical core, which D0.5 freezes and hands to Formalization (middle band). An
  amber review can reject the work or send it back one stage. The four red diamonds
  are halts that call for a judgment from outside the run. D0-max asks an oracle
  whether the result can be sharpened before D0.5 freezes it; ckpt~D/F decides whether
  the result is worth the expense of F1--F5; ckpt~1 and ckpt~2 can ask a person to
  decide, before F2 and before the run is recorded. Most runs used the automatic mode,
  where only ckpt~2 is normally held for a person. Bottom: the substages of the proof
  loop. A green arc sends work back for repair, while an amber arc escalates a false
  or under-specified claim to D0 instead of quietly rewriting the frozen statement.
  \Cref{tab:stage-spec} gives the purpose and acceptance condition of each stage.}
  \label{fig:stage-flow}
\end{figure*}

Here we describe the pipeline of \Cref{sec:pipeline} as it was actually implemented
for the runs we report. A theorem run is named by a question identifier together with a
specialization, and the orchestrator walks it through the ordered stages of
\Cref{fig:stage-flow,tab:stage-spec}. The numbered stages are a finer version of the
four stages of \Cref{sec:pipeline}. Discovery is D-1.1 through D0.5. Formalization is
F1, F1.5, and F2, and proof construction is F3. The statement audit of
\Cref{sec:pipeline-audit} runs at three points of the proof loop, namely F2.5, F3.5,
and F4. F5 prepares the finished record, which the presentation state machine of
\Cref{app:pipeline-paper} then picks up as a separate process. The subsections after
\Cref{tab:stage-spec} describe topic selection and the discovery reviews in more detail.

\begin{longtable}{@{}p{0.11\textwidth}p{0.24\textwidth}p{0.39\textwidth}p{0.18\textwidth}@{}}
  \caption{Stages of a \CausalSmith{} theorem run. The internal state stores the
  numerical identifiers alone; the D and F prefixes are what the command-line
  interface and the logs use to tell discovery apart from formalization.}
  \label{tab:stage-spec} \\
  \toprule
  Stage & Purpose & Action and acceptance condition & Principal durable output \\
  \midrule
  \endfirsthead
  \multicolumn{4}{@{}l}{\small\itshape Table \thetable\ continued} \\
  \toprule
  Stage & Purpose & Action and acceptance condition & Principal durable output \\
  \midrule
  \endhead
  \bottomrule
  \endfoot
  D-1.1 & Literature survey & Searches open problems and earlier proposals for a question worth solving, before anything is proposed. & Gap record \\
  D-1.2 & Proposal construction & Writes a typed proposal together with its mathematical core, which names the objects, the atomic assumptions, the statements, and how they depend on one another. & Proposal and typed core \\
  D-0.5 & Proposal review & Reviews the proposal for novelty, for duplication, and for whether its mathematical direction is viable. If the proposal is rejected, or if a finding looks repairable, it goes back to whichever discovery work can address it, and formalization does not begin. & Review decisions and revision record \\
  D0 & Mathematical derivation & Derives the proposed result and writes the research note. Where the derivation warrants it, the solver may narrow a claim that turns out to be too strong, provided what it hands on for the later \lean{} proof is still a real scientific claim. & Derivation note and updated core \\
  D0-max & Maximality checkpoint & Once the derivation goes through cleanly, halts the run and puts the whole-paper question to an external reviewer: is a sharper bound, a better construction, a stronger reframing, or a tighter constant available? A concrete improvement comes back to D0 as a directive, while a reviewer who confirms the result is already maximal releases the run into D0.5.  & Maximality decision (verbatim reviewer finding logged) \\
  D0.5 & Soundness and novelty review & Keeps a fresh re-derivation of the mathematics separate from the decisions about structure, novelty, and tier. A further cold review is added when the target tier calls for one. & Mathematical, structural, and tier reviews \\
  ckpt~D/F & D0.5$\to$F1 go/no-go & Halts the run once D0.5 has passed. The orchestrator weighs whether a result that is now maximized and cleared for novelty is worth the expense of F1--F5. Resuming enters F1; stopping either records the discovery-only result or downgrades it. & Commit decision and lease re-grant \\
  F1 & Formalization plan & Maps every core node, and the ambient setup it needs, onto a planned \lean{} object. Where a result can be reused, the plan names the declaration and the module it lives in; where a needed fact is unavailable, the plan records an openly disclosed obligation to be discharged. & \code{plan.json} and dependency graph \\
  F1.5 & Plan and reuse review & Runs deterministic checks on coverage, node kind, dependency closure, whether each declaration it cites exists, and where modules are placed, then reviews how well the reuse fits and whether the abstraction level is right. A plan that comes back clean can stop at the first checkpoint so an operator may audit depth, reuse, and statement fidelity; in automatic mode it resumes on its own. & Plan and reuse reviews \\
  F2 & \lean{} scaffold & Writes or revises a \lean{} scaffold. The first scaffold may still contain proof obligations. A later revision touches only the declarations that were flagged and the dependents it has to follow, so proof bodies that already work are left alone. & Tagged \lean{} source tree \\
  F2.5--F4 & Proof and statement review loop & Checks first that the scaffold really does realize the frozen specification (F2.5). The proof-review loop then fills the remaining obligations and rechecks the frontier it changed against the live \lean{} compiler (F3). Next come the deterministic unused-hypothesis lint and the scan for forbidden proof steps (F3.5), and last a full dual-model convergence review of the frozen graph (F4). The numbers F2.5, F3, F3.5, and F4 survive as labels for these substeps, but the work belongs to the loop. & Proof reviews, graph verdicts, crosswalk, and dependency ledger \\
  F5 & Final approval preparation & Begins only once the scans for forbidden proof steps and the correspondence checks come back clean. It then writes a final \TeX{}-to-\lean{} crosswalk covering lemmas as well as theorems, updates the API documentation, and stops at the second checkpoint. Recording, committing, and promotion each need a human to say so. & Complete crosswalk and API update \\
\end{longtable}

\subsection{Topic selection}
\label{app:topic-selection}

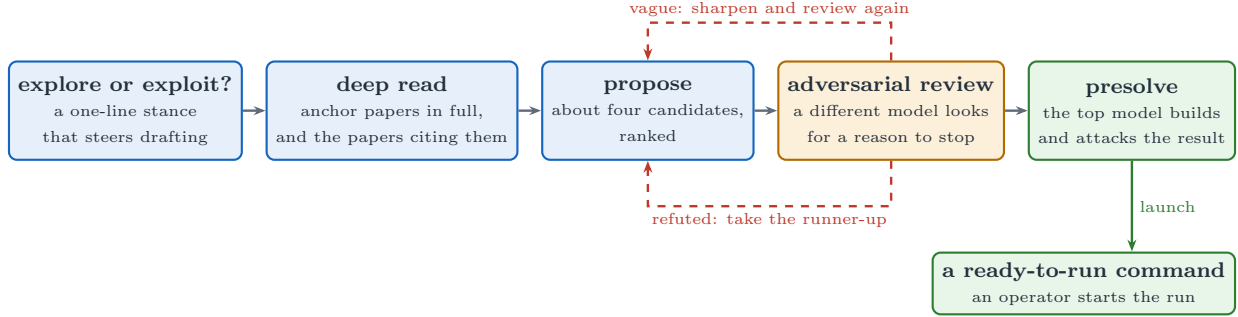
\begin{figure}[!htbp]
  \centering
  \begin{tikzpicture}
    \node[apstage, minimum height=13mm, minimum width=24mm] (mode)
      {\textbf{explore or exploit?}\\{\tiny a one-line stance}\\{\tiny that steers drafting}};
    \node[apstage, right=3mm of mode, minimum height=13mm, minimum width=24mm] (read)
      {\textbf{deep read}\\{\tiny anchor papers in full,}\\{\tiny and the papers citing them}};
    \node[apstage, right=3mm of read, minimum height=13mm, minimum width=24mm] (prop)
      {\textbf{propose}\\{\tiny about four candidates,}\\{\tiny ranked}};
    \node[apcheck, right=3mm of prop, minimum height=13mm, minimum width=24mm] (gate)
      {\textbf{adversarial review}\\{\tiny a different model looks}\\{\tiny for a reason to stop}};
    \node[apgood, right=3mm of gate, minimum height=13mm, minimum width=24mm] (pre)
      {\textbf{presolve}\\{\tiny the top model builds}\\{\tiny and attacks the result}};
    \draw[apar] (mode) -- (read);
    \draw[apar] (read) -- (prop);
    \draw[apar] (prop) -- (gate);
    \draw[apar] (gate) -- (pre);
    \aploopover{(gate.north)}{(prop.north)}{vague: sharpen and review again}
    \aploopunder{(gate.south)}{(prop.south)}{refuted: take the runner-up}
    \node[apgood, below=12mm of pre.south east, anchor=north east, minimum width=40mm] (go)
      {\textbf{a ready-to-run command}\\{\tiny an operator starts the run}};
    \draw[aparg] (pre.south) -- node[aplabg, right=1pt] {launch} (pre.south |- go.north);
  \end{tikzpicture}
  \caption{Topic selection. Blue boxes do the work, the amber box is the adversarial
  review, and red dashed arrows send a candidate back. Only a launch verdict from the
  presolve produces a command.}
  \label{fig:topic-selection}
\end{figure}

A run invoked with no topic starts with the topic-selection module, ahead of the numbered
D stages. If a researcher supplies the topic instead, the module is skipped and the run
enters D-1.1 directly. \Cref{fig:topic-selection} shows the flow.

The module first states a stance in one line. An exploit stance extends a recorded
result, or works in a cluster where many attempts have been accepted. An explore stance
goes to an area with few acceptances. The stance shapes how candidates are drafted, and
every candidate is then ranked and reviewed on the same terms.

The module then searches recent work in the area and picks one or two anchor papers
that carry a theorem. Subagents read each anchor in full, together with the papers that
cite it. Each read returns a short digest: the central theorems and their load-bearing
proof steps, one or two questions left open at the boundary of the proof, and one or
two directions where the same technique could reach a new target or setting. From
these digests the module drafts about four candidates. It ranks them by evidence,
mathematical depth, novelty, feasibility, and value to a named consumer, that is, a
published applied work or a software default whose practice would change if the result
holds. Each candidate is also checked against the runs that are already active or
recorded.

The top candidate goes to an adversarial review run on a different model. The reviewer
looks for a concrete reason to stop: a counterexample, a reduction to something
trivial, or a paper that already has the result. It also asks whether the conjecture is
stated precisely, whether the result would clear its target tier if true, and whether
the consumer is real. A vague candidate comes back to be sharpened and is reviewed
again, up to three times. A refuted candidate is dropped and the runner-up takes its
place. If both fall, the module drafts one fresh slate, with the rejected topics and
their failure modes as constraints.

A candidate that passes review goes to a presolve. In a single call the strongest
available model attempts the result, attacks its own attempt with counterexamples and a
fresh literature search, and writes the strongest supported theorem at full precision,
with every gap listed. The call returns launch, pivot, or drop. Learning at this point
that a result is out of reach costs one model call, while learning it during derivation
costs a day of pipeline time. A launch verdict produces a ready-to-run command whose
topic string carries a short summary of the presolve and the path to its draft, and an
operator starts the run from it. Discovery treats the draft as evidence that still has
to be verified.

\subsection{Proposal review}
\label{app:proposal-review}

D-0.5 reviews the proposal against a rubric written in advance. A mechanical check has
already confirmed that the proposal is well formed, so the reviewer spends its effort on
meaning. Every finding carries one code from one of three families, and the family
decides where the finding goes.

Structure findings check that the strongest result of the proposal is its headline
target, instead of a lemma on the way to a weaker theorem. They lead to a structural
patch.

Novelty findings are judged per statement on two axes, our own past runs and the
published literature, and a statement counts as new only when it is new on both. A
finding that a result is already known names the run or the paper. A finding that the
proposal misstates a cited paper, or that the result is already published, has to come
with a receipt giving the exact source version and a page, section, or equation. These
findings send the proposal back for a redraft or a pivot.

Soundness findings ask whether each statement is well defined, whether it reduces
correctly in the special cases it claims, and whether each assumption tagged as standard
really is the standard condition under that name. At this stage soundness concerns the
question itself; the truth of each statement is settled later, in D0. These findings
lead to a redraft of the statement.

The review closes with a tier rating, scored against the target tier fixed when the run
was launched. A repairable proposal returns to drafting a bounded number of times. A
proposal that is already known, or too thin for its tier, is recorded and dropped.

\subsection{Solving and derivation review}
\label{app:solve-review}

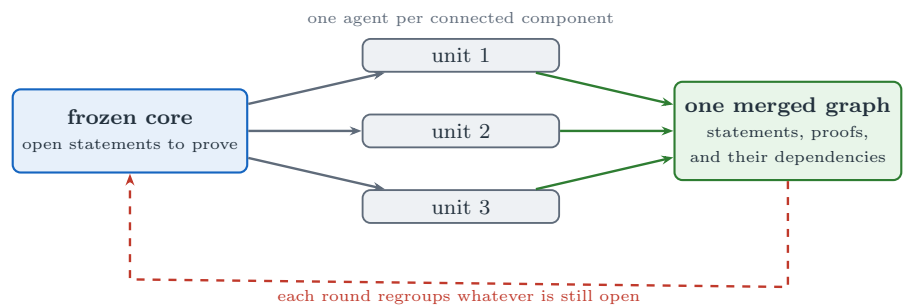
\begin{figure}[!htbp]
  \centering
  \begin{tikzpicture}
    \node[apstage, minimum width=28mm, minimum height=11mm] (q)
      {\textbf{frozen core}\\{\tiny open statements to prove}};
    \node[apquiet, right=15mm of q, yshift=10mm, minimum width=26mm] (u1) {unit 1};
    \node[apquiet, right=15mm of q, minimum width=26mm]               (u2) {unit 2};
    \node[apquiet, right=15mm of q, yshift=-10mm, minimum width=26mm] (u3) {unit 3};
    \node[aplab, above=1mm of u1] {one agent per connected component};
    \foreach \n in {u1,u2,u3} { \draw[apar] (q) -- (\n); }
    \node[apgood, right=15mm of u2, minimum width=30mm, minimum height=13mm] (g)
      {\textbf{one merged graph}\\{\tiny statements, proofs,}\\{\tiny and their dependencies}};
    \foreach \n in {u1,u2,u3} { \draw[aparg] (\n) -- (g); }
    \aploopunder[14mm]{(g.south)}{(q.south)}{each round regroups whatever is still open}
  \end{tikzpicture}
  \caption{Solving at D0. The open statements are split by the connected components of
  their dependency graph, the units are solved concurrently, and their outputs merge
  back into one graph.}
  \label{fig:d0-units}
\end{figure}

D0 solves the frozen proposal core, which already fixes the statements to prove and the
dependencies between them. The solver groups the open statements into units (\Cref{fig:d0-units}), one per
connected component of their dependency graph, so two open statements share a unit when
a chain of open dependencies joins them. Each unit goes to its own agent, and the units
run concurrently. An agent owns only the targets of its unit. It may add helper lemmas
with their own proofs, and it declares every lemma and assumption its proofs rely on, so
that the outputs of all units merge into one graph. A classical published theorem enters
as a cited leaf, carried with its source and its exact statement. Each round regroups
whatever is still open.

When a claim turns out to be too strong, the solver weakens it to something true and
still interesting, or corrects the definition behind it. When a claim turns out to be
too weak, the solver strengthens it. Either way the changed statement goes through the
reviews again.

\begin{figure}[!htbp]
  \centering
  \begin{tikzpicture}
    \node[apgood, minimum width=24mm, minimum height=12mm] (g)
      {\textbf{derived note}\\{\tiny statements and proofs,}\\{\tiny as a graph}};
    \node[apcheck, right=13mm of g, yshift=18mm, text width=64mm, align=left] (m)
      {\textbf{math reviewer}: does the derivation hold?\\
       {\tiny re-derives each load-bearing step from the definitions,}\\
       {\tiny then compares its derivation with the note term by term}};
    \node[apcheck, right=13mm of g, text width=64mm, align=left] (r)
      {\textbf{rubric reviewer}: does it clear the written bar?\\
       {\tiny structure, novelty, and positioning}};
    \node[apcheck, right=13mm of g, yshift=-18mm, text width=64mm, align=left] (n)
      {\textbf{general referee}: would a referee care?\\
       {\tiny a cold read with plain tier definitions and no rubric,}\\
       {\tiny calibrated on scored accepted papers; its tier caps the result}};
    \foreach \x in {m,r,n} { \draw[apar] (g) -- (\x); }
    \node[apgood, right=10mm of r, minimum width=19mm] (go)
      {\textbf{go}\\{\tiny to F1}};
    \draw[aparg] (m.east) to[out=0,in=130] (go.north west);
    \draw[aparg] (r.east) -- (go.west);
    \draw[aparg] (n.east) to[out=0,in=-130] (go.south west);
  \end{tikzpicture}
  \caption{The derivation review at D0.5. Each referee answers one question. The general
  referee joins when the target tier calls for it. A math failure, or a salvageable
  result below the target tier, sends the run back to D0.}
  \label{fig:d05-referees}
\end{figure}
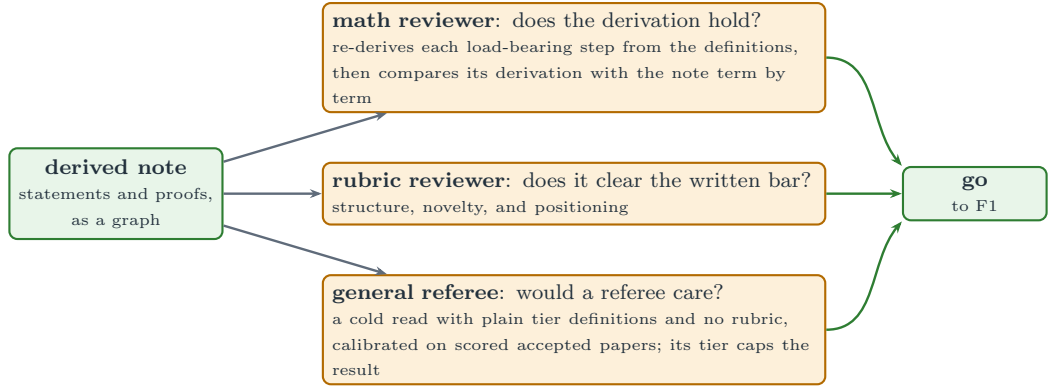

D0.5 splits the review of the derivation across separate referees
(\Cref{fig:d05-referees}). Correctness, fit with
the rubric, and interest pull in different directions, and a single reviewer asked all
three tends to pass a correct but dull result on correctness alone. The math reviewer
re-derives each load-bearing step from the definitions, then compares its own
derivation with the note term by term and reports every difference. The rubric reviewer
checks structure, novelty, and positioning against the written rubric. When the target
tier calls for it, a general referee reads the note cold. It sees the note and plain
tier definitions and nothing else. The solver itself works toward the rubric, so a note
can satisfy every item and still fall short, and the general referee is kept away from
the rubric for that reason. It grades what is proved over what is promised, and it
calibrates against a spread of accepted papers together with the scores their final
referee gave them.

A failure from the math reviewer sends the run back to D0. A general referee that places
the result below the target tier, but judges it salvageable, sends the run back to D0
with a directed improvement, at most twice by default. A result judged below the target
tier and beyond repair halts the run for a decision.

\section{How the graph drives formalization, proof, and audit}
\label{app:pipeline-proof-loop}

The pipeline is built around the graph. The D stages create it in natural language.
The typed core of D-1.2 names the objects, assumptions, and statements together with the
edges between them, D0 extends it while deriving the result, and D0.5 freezes the
argument, one node per claim. The F stages attach \lean{} to it. F1 gives every node an
intended declaration and module, F2 emits the scaffold, and each pass of the proof loop
reads the emitted declarations and dependencies back into the graph. Formalization also
adds nodes for the intermediate lemmas the derivation never had to name. The graph thus
maps each planned object to what it should become, and each emitted declaration to what
it realizes.

At F1.5, code checks what has a definite answer: whether each cited declaration exists,
whether every node is covered, and whether the dependencies close. The model judges the
rest. It asks whether a reused lemma sits at the right level of abstraction, whether the
plan searched the library before rebuilding, and whether a missing fact should be built
or carried as an openly disclosed assumption.

The statement audit runs at F2.5, F3.5, and F4; a second audit, in Presentation, checks
the paper against the frozen graph (\Cref{app:pipeline-paper}). Each node is
\code{unreviewed}, \code{matched}, or \code{drift}. An accepted review is stamped with a
fingerprint of the content it reviewed, so a change to a node or its realization sends
that node and its downstream nodes back to review.

In the statement phase, the reviewer compares each frozen statement and definition with
the \lean{} scaffold. Typical drift is a dropped hypothesis, a constant fixed where the
claim was general, or a definition narrowed until the claim becomes easy. A mechanical
mismatch goes back to F2 for a local revision. A false or under-specified claim, or a
missing library fact, is escalated. The frozen natural-language statement stays as
written throughout, since it is the contract the audit protects.

Proof filling starts once the statement phase is clean. One prover session runs at a
time, because the statements of one result sit in a single import chain, and each
session starts from a summary of the earlier ones. Each iteration reviews the frontier
it changed, attempts the open obligations against the live compiler, refreshes the
graph, and checks for progress. The loop finishes with a successful build, no
\code{sorry} or \code{admit}, the frozen theorems and their dependency closures proved
and matched, a clean scan for \code{axiom}, \code{opaque}, and \code{unsafe}, and a
passing unused-hypothesis lint. An unused hypothesis suggests that the assumption is
decoration or that the statement says less than intended. A blocking finding returns
the loop to the statement phase, and the rest are recorded for the orchestrator.

The final convergence review runs independently of the incremental one, on GPT and on
Opus. It revisits every object that originates in the paper, each theorem, definition,
lemma, and symbol with its assumptions, and judges helper lemmas through the statements
that use them. Two different models share fewer blind spots than one model asked twice.
When they disagree, the orchestrator reproduces the complaint. A correct complaint is
fixed in the code, and a mistaken one becomes a general fix to the reviewer prompt. The
review also records how each library dependency was treated. A \statusgated{} dependency
appears as an explicit conditional assumption or as a visible build obligation. A
\statuscited{} dependency is checked against its source and blocks approval if its
encoding is mismatched or underspecified. F5 then writes the complete crosswalk, updates
the API documentation with a doc comment on every public declaration, and records the
run as accepted, downgraded, or failed.

\section{Presentation pipeline}
\label{app:pipeline-paper}

\begin{figure}[!htbp]
  \centering
  \begin{tikzpicture}[ps/.style={minimum width=19mm, minimum height=15mm}]
    \node[apstage, ps] (p0) {\textbf{P0}\\literature\\{\tiny verify each}\\{\tiny bibliography entry}};
    \node[apstage, ps, right=2.5mm of p0] (p1) {\textbf{P1}\\render\\{\tiny each declaration,}\\{\tiny then freeze it}};
    \node[apstage, ps, right=2.5mm of p1] (p2) {\textbf{P2}\\write\\{\tiny prose, proofs,}\\{\tiny related work}};
    \node[apcheck, ps, right=2.5mm of p2] (p3) {\textbf{P3}\\check\\{\tiny code checks and}\\{\tiny a prose rubric}};
    \node[apstage, ps, right=2.5mm of p3] (p4) {\textbf{P4}\\emit\\{\tiny build, and link}\\{\tiny claims to code}};
    \node[apcheck, ps, right=2.5mm of p4] (p5) {\textbf{P5}\\referee\\{\tiny read it as a}\\{\tiny journal would}};
    \node[apgood, ps, right=2.5mm of p5] (p6) {\textbf{P6}\\slides\\{\tiny optional deck}\\{\tiny from the paper}};
    \foreach \a/\b in {p0/p1, p1/p2, p2/p3, p3/p4, p4/p5, p5/p6} { \draw[apar] (\a) -- (\b); }
    \aploopunder{(p5.south)}{(p1.south)}{each finding returns to the earliest stage that can act on it}
  \end{tikzpicture}
  \caption{Stages of the presentation pipeline. Blue boxes produce the paper, amber boxes
  review it, and the red dashed arrow is the revision route taken after the referee
  review.}
  \label{fig:paper-stages}
\end{figure}
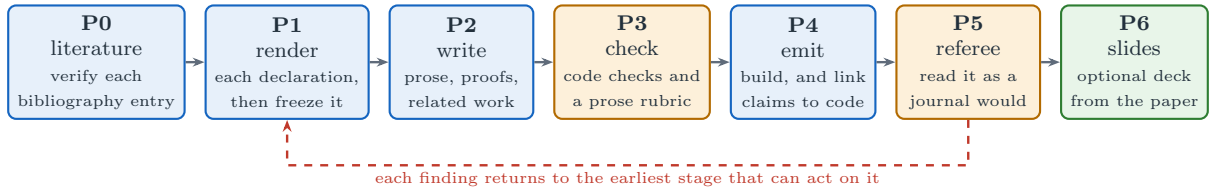

The theorem pipeline and the paper pipeline are separate state machines. The paper
pipeline takes a recorded result and runs stages P0 through P5, with an optional P6 for
slides (\Cref{fig:paper-stages}). The paper is generated from the checked code, and
every later pass asks whether the prose has drifted from it.

P0 assembles the literature. A model proposes the bibliography, and code looks up each
entry in Crossref, arXiv, and OpenAlex and drops any entry that fails to resolve. Every
citation in the paper has to come from this verified pool.

P1 builds the formal layer of the paper. It walks the graph and renders each verified
declaration as a displayed theorem, definition, or assumption in ordinary mathematical
notation, ordered so that every object appears before anything that uses it. Each
rendered statement is judged against its \lean{} declaration at the moment it is
rendered. The English statement and its \lean{} counterpart have to be logically
equivalent in assumptions and in conclusion alike. If either one is the stronger, the
pipeline halts for adjudication and keeps the declaration it was aiming at. The
accepted layer is then frozen. Later passes rewrite the prose around it freely, and
every frozen statement is restored byte for byte afterwards.

P2 writes everything else: the introduction and abstract, the motivation, the proof of
each result, and the comparison with related work. The prose that describes proofs is
audited against the graph. When that audit fails, the pipeline runs one promotion
round. It adds the statements the proofs need to the formal layer, renders them at P1,
and retries P2 once.

P3 reads the assembled draft as one document. Code checks first that every citation key
lies in the verified pool, that every cross-reference resolves, and that the statements
stand in the order P1 fixed. A model then asks whether any sentence claims more than the
theorems prove, whether each cited source supports the sentence that cites it, and how
the draft scores on a rubric covering claims, positioning, assumptions, and writing.
Revisions touch the prose alone, and the draft gets at most two rounds.

P4 emits the rendered paper with its links to the code. Links are assembled from the
graph first. For each object it displays, the presentation system pulls the relevant
declaration and its \code{statement-uses} neighbors out of the verified graph, emits a
source anchor, and checks that the dependencies and assumptions on display have links
of their own. A link therefore resolves to an object that has passed the statement
audit. P4 also ties each prose statement to its declaration hypothesis by hypothesis,
which drives the crosswalk of \Cref{app:site}.

P5 reads the paper as a journal referee would, taking the theorems as given. It asks
whether the contribution is significant and delivered, whether each prose claim matches
the verified statement in strength and scope, whether the paper engages the actual
results of its closest competitors, and whether each condition is stated where it is
claimed. Each finding is routed to the earliest stage that can act on it. The
orchestrator then revises the sources by hand and re-enters the pipeline at that stage,
after which the downstream reviews run again. P1 and P2 double as outline and draft
checkpoints: resuming from either one approves going on, and re-entering explicitly at
P0, P1, or P2 allows a targeted revision. Once the paper settles, P6 can build the slide
deck, taking every statement verbatim from the frozen layer.

\section{Study mode}
\label{app:study}

Partway through a proof, formalization can reach a step that needs a lemma nobody has
written. Three responses are reasonable. The run can record the gap and ship as an
openly conditional result, leaving the lemma for a later run. It can prove the lemma in
place, next to the paper, which is the quickest route to a finished result. Or it can
build the lemma for the library, stated at its natural generality, so that later runs
find it ready. The choice turns on how wide the gap is and how many results are likely
to want the same fact. The orchestrator sends a fact of general use to study mode, and
builds a fact that is specific to one model and of manageable size in a helper file
inside the run. The theorem run carries on with the fact assumed. Once the lemma lands,
the new declaration returns to F1 as a candidate for reuse, and the statements that use
it are reviewed again.

\begin{figure}[!htbp]
  \centering
  \begin{tikzpicture}
    \node[apbox, minimum width=25mm, minimum height=14mm, align=left] (req)
      {\textbf{a written requirement}\\{\tiny the fact, its generality,}\\{\tiny how the literature states it}};
    \node[apstage, right=5mm of req, minimum width=31mm, minimum height=14mm] (plan)
      {\textbf{plan and scaffold}\\{\tiny which lemmas, in what order;}\\{\tiny statements written, proofs open}};
    \node[apstage, right=5mm of plan, minimum width=26mm, minimum height=14mm] (fill)
      {\textbf{prover}\\{\tiny one at a time, against}\\{\tiny the live compiler}};
    \node[apcheck, right=5mm of fill, minimum width=23mm, minimum height=14mm] (rev)
      {\textbf{reviewer}\\{\tiny seven checks,}\\{\tiny confirmed by the build}};
    \foreach \a/\b in {req/plan, plan/fill, fill/rev} { \draw[apar] (\a) -- (\b); }
    \aploopover{(rev.north)}{(plan.north)}{rework, up to three rounds}

    \node[apgood, below=10mm of fill, minimum width=36mm, minimum height=11mm] (coord)
      {\textbf{coordinator}\\{\tiny reuses what exists, picks the}\\{\tiny narrowest home, writes docstrings}};
    \draw[aparg] (rev.south) |- (coord.east);
    \node[apquiet, left=6mm of coord, minimum width=34mm, minimum height=11mm] (gate)
      {\textbf{integration gate}\\{\tiny build, index, embeddings,}\\{\tiny documentation checks}};
    \draw[apar] (coord) -- (gate);
    \node[apgood, left=10mm of gate, minimum width=22mm, minimum height=11mm] (out)
      {in \Causalean{}};
    \draw[aparg] (gate) -- node[aplabg, above=1pt] {pass} (out);
    \aploopunder{(gate.south)}{(coord.south)}{fail: roll back and retry, up to three times}
  \end{tikzpicture}
  \caption{A study-mode build. The top row writes and proves the module; the bottom row
  places it in \Causalean{}. Red dashed arrows are the bounded rework routes.}
  \label{fig:study-build}
\end{figure}
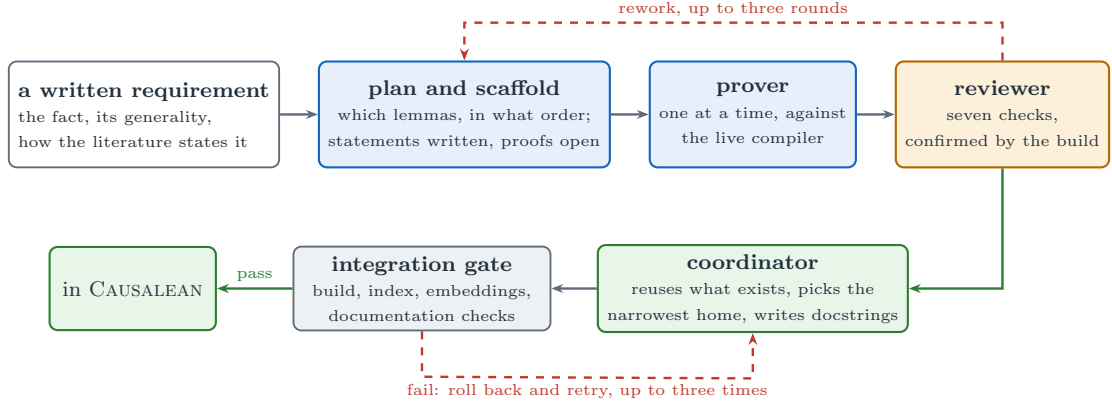

A study starts from a short requirement written in plain English
(\Cref{fig:study-build}). It states the fact
that is needed, the generality it should have, and how the literature states it. The
requirement leaves the home of the code open. A scaffolder plans the lemmas and their
order and writes a file with every statement present and every proof left open. A prover
then fills the proofs against the live compiler. Provers run one at a time for the same
reason as in F3, since the new files usually form one import chain. After each round the
scaffolder reads the build report and decides whether to run another round or send the
module to review, with at most ten build rounds.

A reviewer then judges the module against seven checks, all of which have to hold. The
module is generic: it states the general fact, at the level where the fact holds. It is
reusable: a clean, well-named statement whose hypotheses are the ones a later proof
would reach for. It is standard: it matches how the literature states the result. It is
non-vacuous: its hypotheses are satisfiable by real inputs, and the statement keeps its
full strength on the hard case. It fulfills the requirement, with sharp constants
intact; the reviewer judges the result delivered and accepts a proof route that differs
from the one the requirement sketched. It is free of \code{sorry}. And it is layered:
its imports stay within Mathlib, \Causalean{}, and the files of the study itself. The
last check protects the direction of dependency, since a library lemma that imports a
research file would carry that paper into the library for good. The build confirms a
passing verdict on its own, so a module that fails to compile or still has an open proof
goes back to the scaffolder. The review allows three rounds.

A coordinator then places the module. It searches \Causalean{} for declarations that
already prove the same statement and reuses them. It picks the narrowest existing file
or subject area that fits, and merges into an existing file where it can. Its edits are
additive, consisting of new files and inserted blocks, so every existing line of the
library stays as it was. It also writes the docstrings that the library explorer and
the retrieval index read. Code then runs the integration gate: a full build,
re-indexing, the retrieval embeddings, and the documentation checks. A failure rolls
\Causalean{} back to its prior state, and the coordinator retries with the failure log,
up to three times. A gate that hangs stops for the orchestrator, since rolling back in
the middle of a build is unsafe. Concurrent studies take turns at this step, because
each one edits the shared library. Once the gate passes, the lemma is part of
\Causalean{}. The library therefore grows out of what the pipeline demands, with a
boundary one can audit between a result specific to a single run and a theorem the
library will reuse.

\section{The reading interface}
\label{app:site}

A finished run leaves three things behind: a paper, a \lean{} development, and the
frozen graph that ties them together. The companion site at
\url{https://jiyuan-tan.github.io/CausalSmith/} presents all three on one page. It adds
four things on top of the text. It draws the dependency structure of the paper's own
proofs. It puts each English statement next to the \lean{} declaration that was audited
against it. It lets a reader record what they checked and what they thought of it. And
it presents the same result as a short 20mins talk. The screenshots in this section all come
from one accepted run, ``Sharp Minimax Rates for Average Treatment Effects with
Discrete Confounding under Fixed Overlap'', and show the site as it stands. 

\subsection{The paper page}

\begin{figure}[!htbp]
  \centering
  \includegraphics[width=\textwidth]{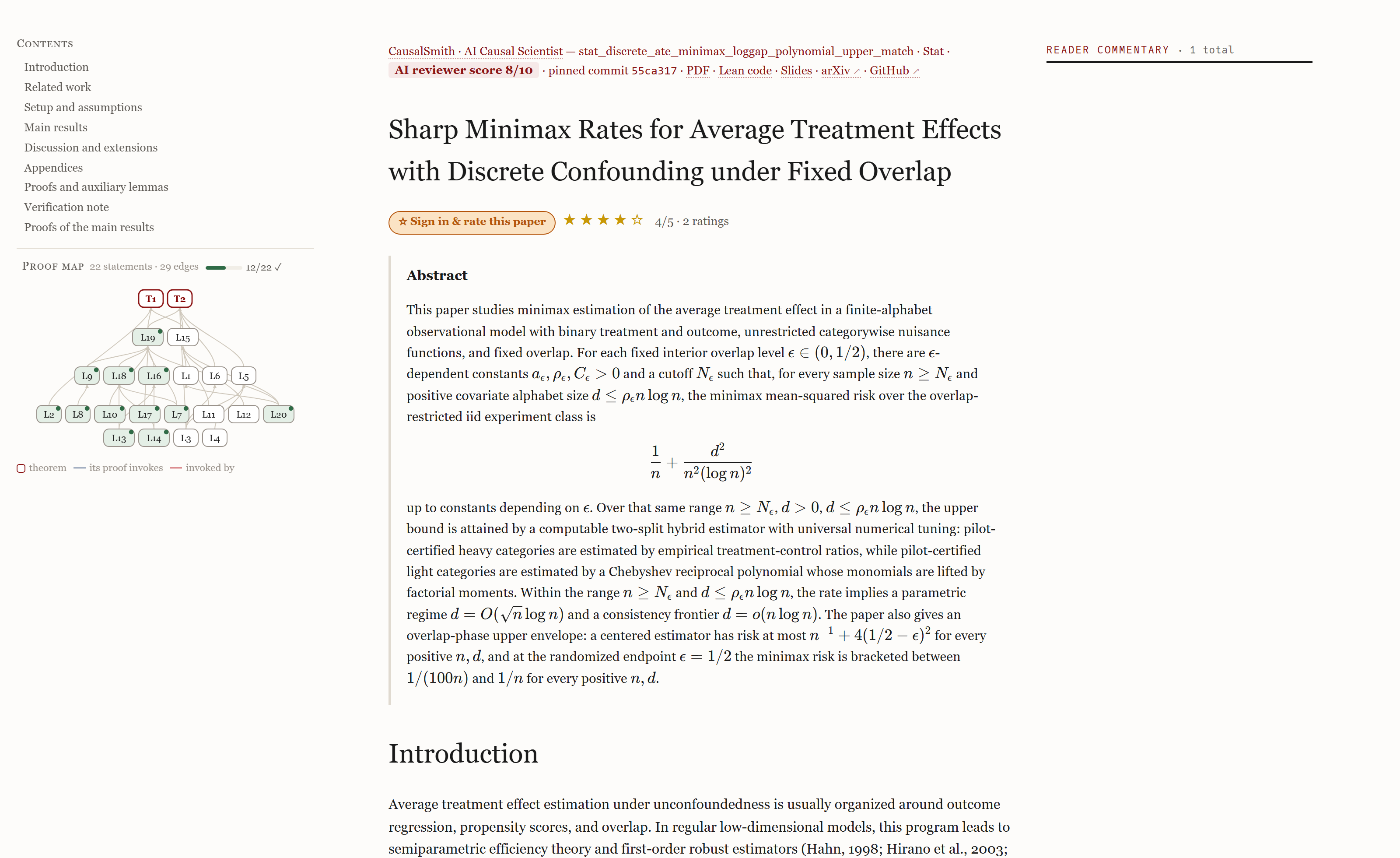}
  \caption{A paper page. Contents and proof map on the left, the paper in the middle,
  reader comments on the right. The byline gives the reviewer score, the reader rating
  (four of five stars from two readers), the pinned commit, and links to the PDF, the
  \lean{} code, and the slides.}
  \label{fig:ui-paper}
\end{figure}

\Cref{fig:ui-paper} shows a paper page. The middle column is the paper itself. The left
rail holds the table of contents and the proof map. The right rail holds reader
comments, each one beside the passage it is about. The byline carries the pipeline's own
reviewer score, the reader rating, the commit version number the page is pinned to, and links to the
PDF, the \lean{} development, and the slides. 

\subsection{Proof map}

\begin{figure}[!htbp]
  \centering
  \includegraphics[width=\textwidth]{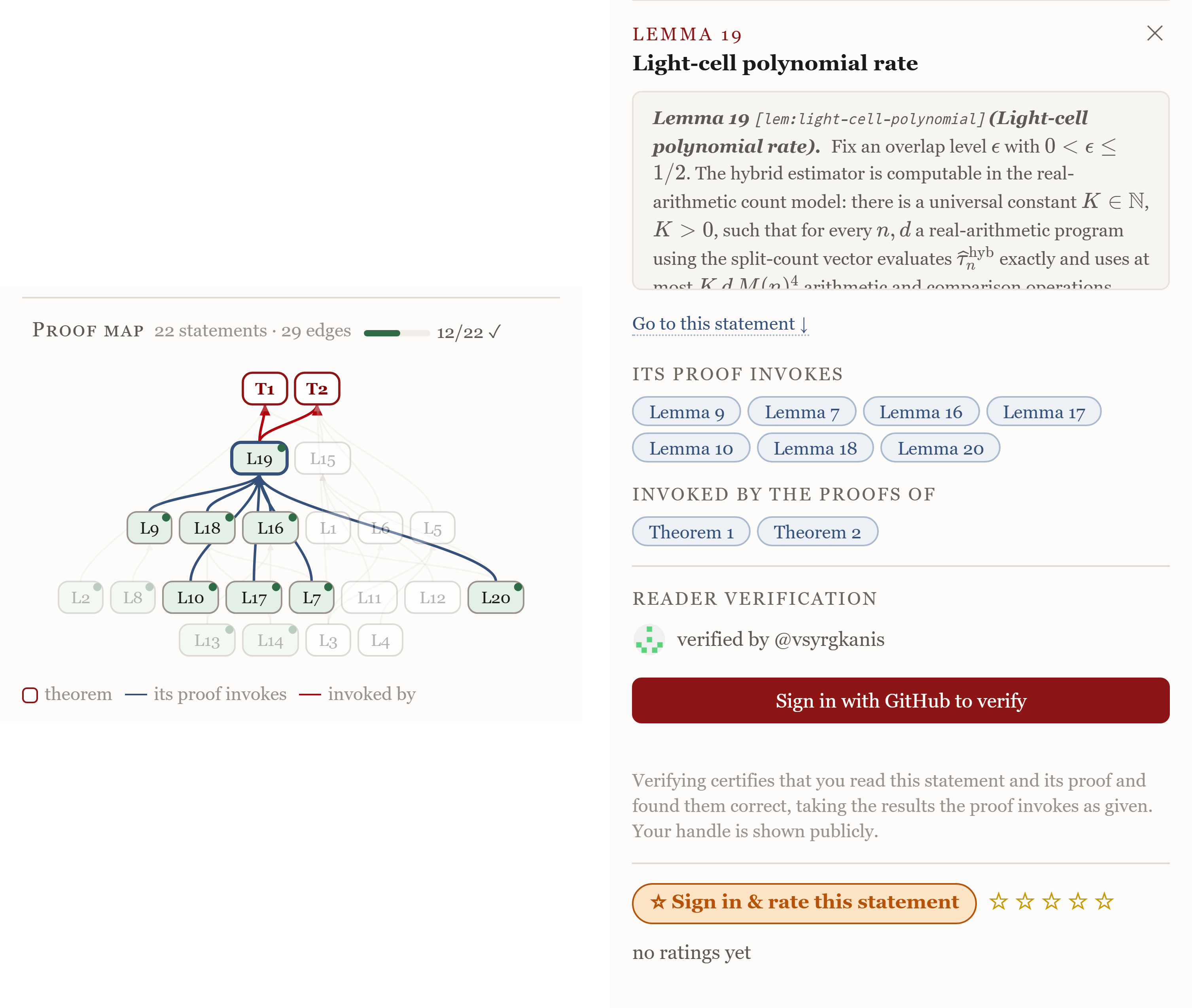}
  \caption{The proof map, with one lemma selected. Left: the map itself, with the
  selected node's dependencies drawn in and the rest of the graph faded. Blue edges run
  to the results the selected proof invokes; red edges run to the two theorems whose
  proofs invoke it. Right: the card the selection fills, giving the statement, both
  dependency lists as links, and the reader marks. The header counts how many of the 22
  statements a reader has verified.}
  \label{fig:ui-proofmap}
\end{figure}

The proof map (\Cref{fig:ui-proofmap}) is the reader-facing view of the frozen graph of
\Cref{fig:graph}, restricted to the results the paper displays. Each node is a theorem,
proposition, or lemma of the paper. An edge means that one proof invokes the other. The
layout is layered: a result sits one rank above the deepest result its proof invokes,
and the theorems are pulled to the top row, so the paper's headline results read as the
roof of the structure. The paper shown here has 22 statements and 29 such edges.

Selecting a node highlights the results its proof invokes and the results that invoke
it. It also fills a card underneath with the statement, both lists as links, and the
reader marks on that statement. Every node is also a link, so a reader can go straight
from the map to the statement it names.

\subsection{Natural-language statement and \lean{} declaration}

\begin{figure}[!htbp]
  \centering
  \includegraphics[width=\textwidth]{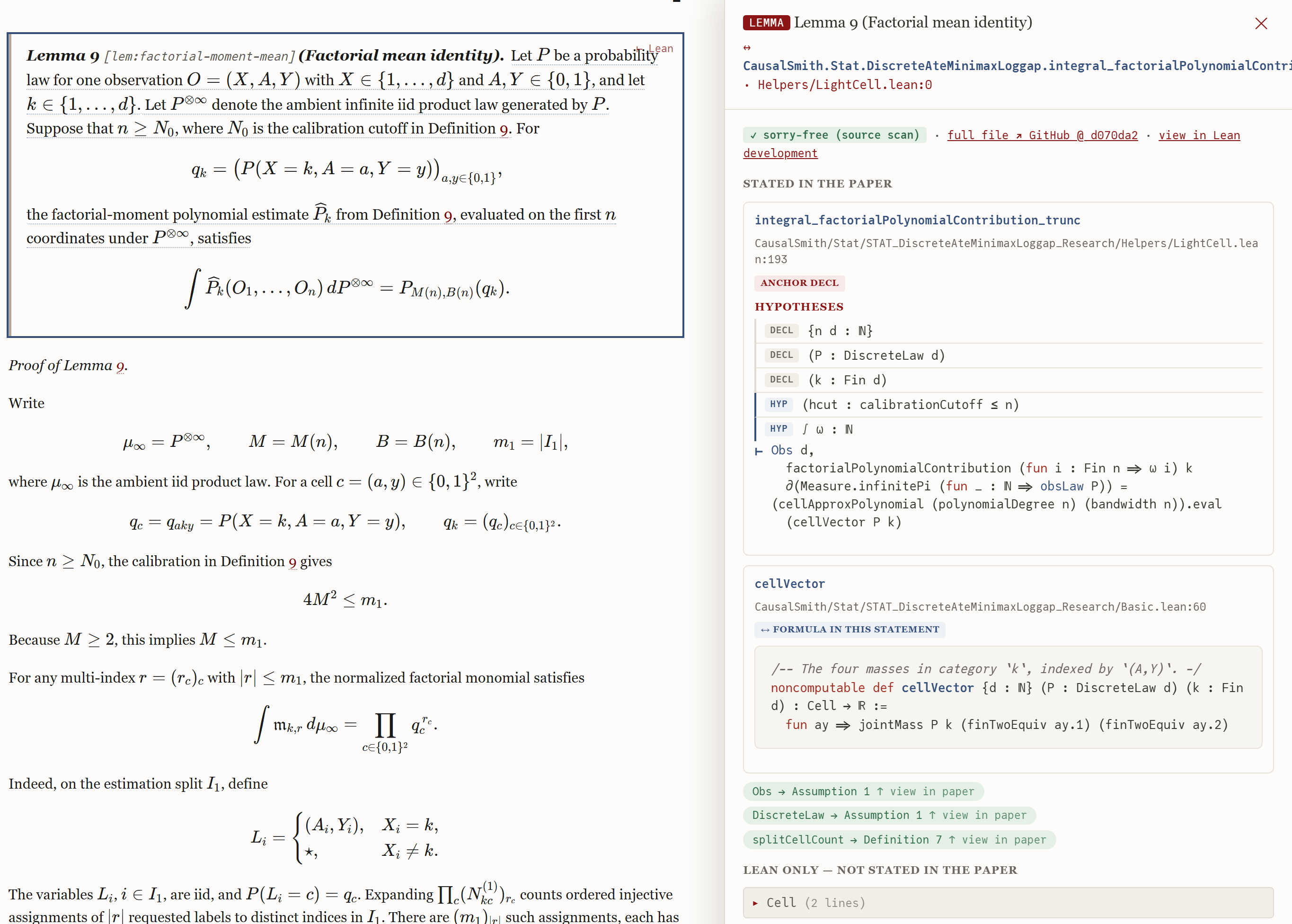}
  \caption{A statement in the paper (left, outlined) next to its \lean{} declaration
  (right). The panel gives the source location, the \code{sorry} scan, the hypotheses
  and the conclusion, and the definitions the statement rests on, each linked back to
  the paper. The last rows list declarations the \lean{} development uses and the paper
  leaves unstated.}
  \label{fig:ui-crosswalk}
\end{figure}

Every statement in the paper is clickable, and clicking it opens the \lean{} side of it
(\Cref{fig:ui-crosswalk}). The panel gives the declaration name, its file and line,
whether a scan of that source finds it free of \code{sorry}, and the statement itself,
split into its hypotheses and its conclusions. Where a symbol in the paper is realized
by a \lean{} definition, that definition is shown too, with a link back to the place in
the paper that introduces it. Two kinds of gap carry an explicit label: a hypothesis the
\lean{} statement holds that the paper leaves unstated, and a declaration in the
development that no paper statement cites.

The pipeline claims that the English statement and the \lean{} one express the same proposition. The
crosswalk puts the two side by side, so a reader can make the same comparison and
disagree with it. A separate page lists the complete development for the paper, module
by module, including the helper lemmas the paper never cites.

\subsection{Reader Interaction}

A reader signed in with a GitHub account can leave three kinds of mark. All are public
and all are attributed.

The first is a verification of a single statement, in the site's own words: I read this
statement and its proof and found them correct, taking the results the proof invokes as
given. The unit is deliberately one statement. The results a proof invokes are somebody
else's to check, so the work is small enough that a reader can finish it in one sitting.
The proof map shows how far this has gone, both as a count in its header and as a dot on
each node. The paper in \Cref{fig:ui-proofmap} has 12 of its 22 statements verified.

The second is a rating of one to five stars, on the paper as a whole or on a single
statement, one per account and withdrawable by clicking the same star again. The page
shows the average and the count. This sits beside the pipeline's own reviewer score and
answers a different question. The score is one model's judgment at the end of the run.
The rating is what readers thought afterwards. The paper of \Cref{fig:ui-paper} has four
of five stars from two ratings.

\begin{figure}[!htbp]
  \centering
  \includegraphics[width=\textwidth]{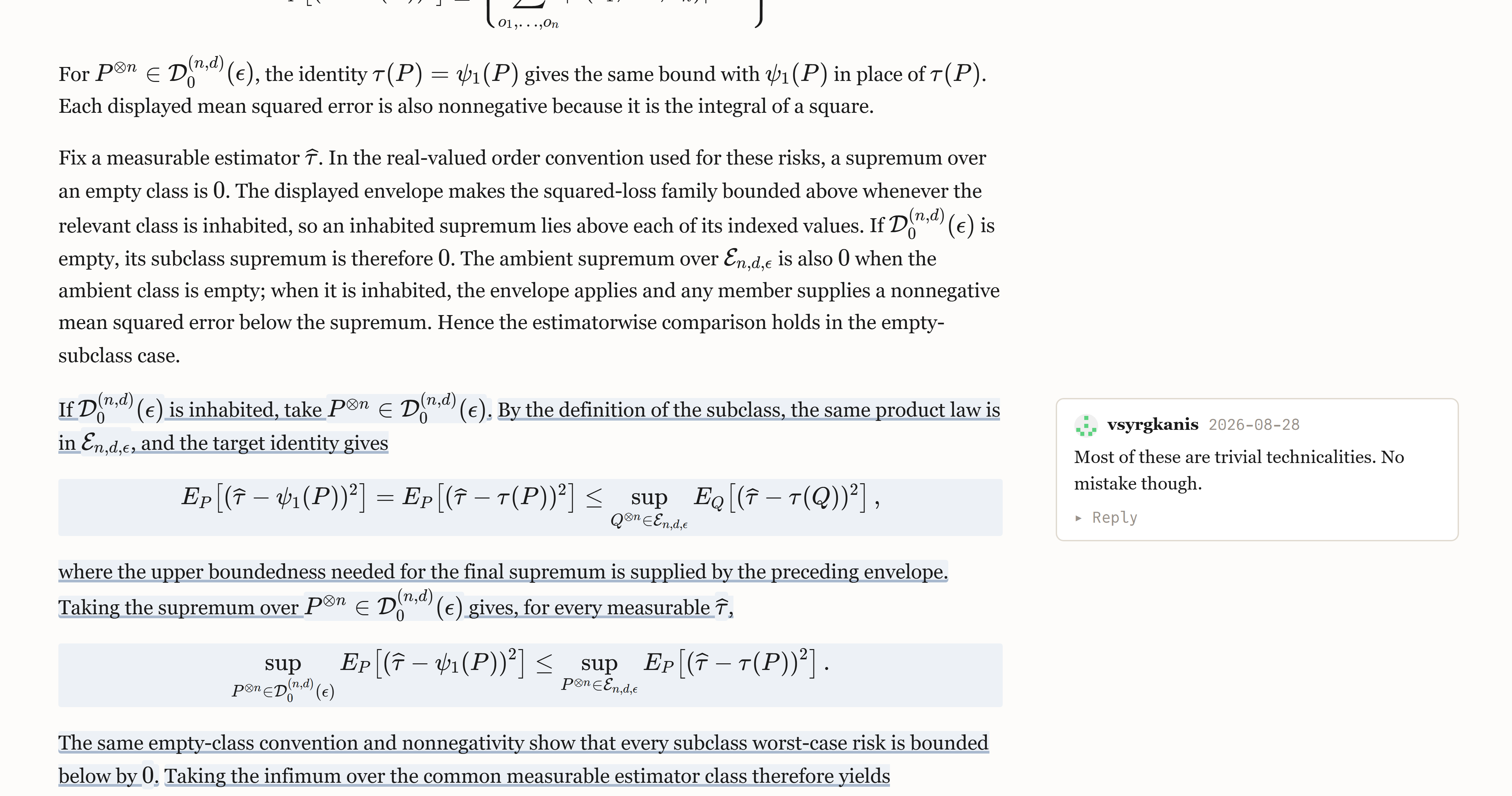}
  \caption{A reader comment. The quoted passage is shaded in the proof and the comment
  sits beside it in the margin, with its author, its date, and a reply thread.}
  \label{fig:ui-comment}
\end{figure}

Selecting any passage of the paper offers to comment on it, and the comment is filed in
the margin beside the passage it quotes (\Cref{fig:ui-comment}). A comment can be left
plain, or tagged as a verification or as a problem, and replies thread underneath it.

\subsection{Slides}

\begin{figure}[!htbp]
  \centering
  \includegraphics[width=\textwidth]{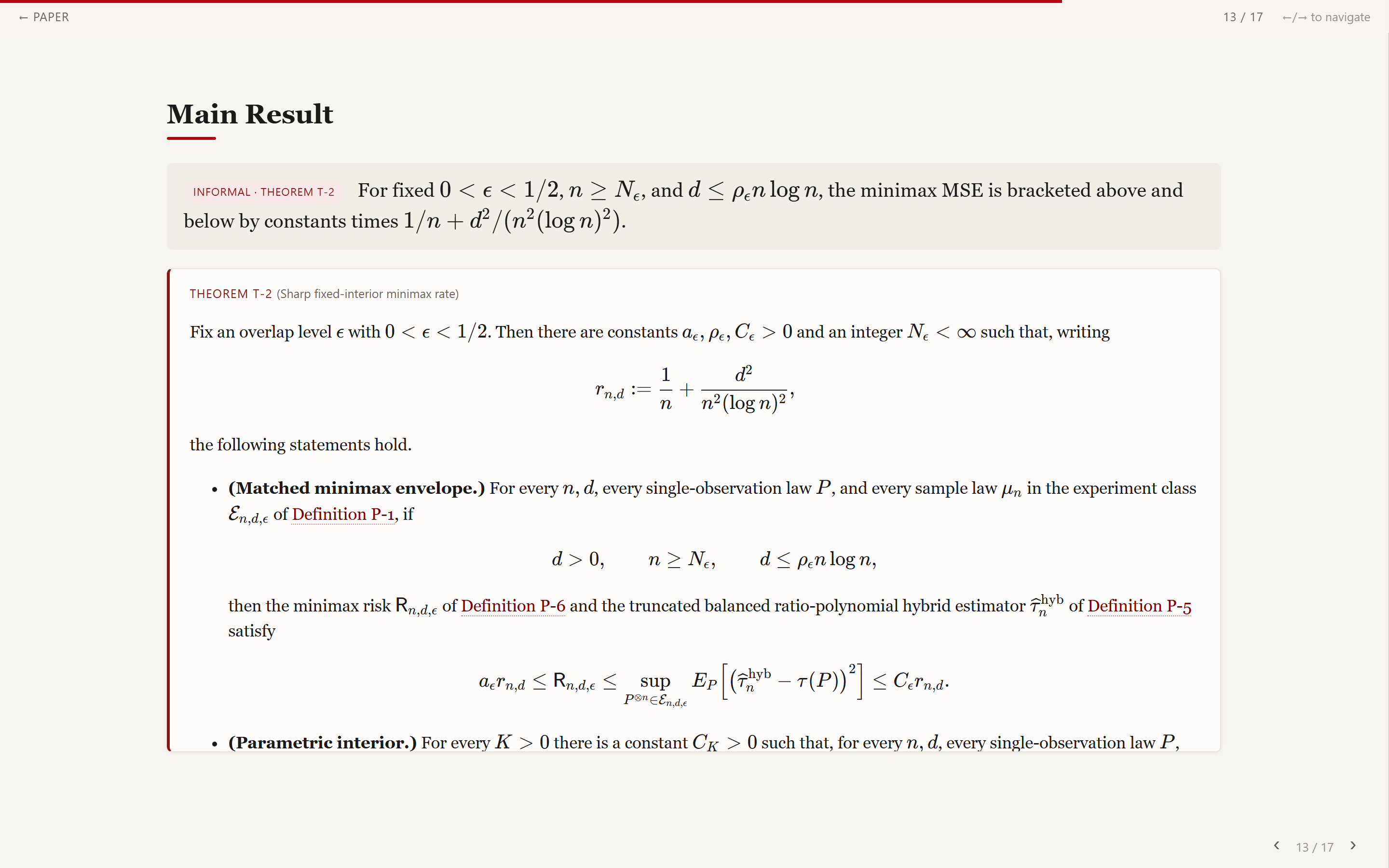}
  \caption{A slide carrying the main result. The one-sentence restatement at the top is
  labeled informal, and the audited statement follows below it, with its definitions
  linked back to the paper.}
  \label{fig:ui-slides}
\end{figure}

Each accepted result also ships as a short deck, built from the same graph as the paper
(\Cref{fig:ui-slides}). The deck for this paper runs to 17 slides: the setting, the
literature position, the idea, the main result, a proof sketch, and what follows.

\clearpage

\section{Reproducibility}
\label{app:repro}

The library and the pipeline are both pinned to \lean{} toolchain
\code{leanprover/lean4:v4.29.0-rc3}. We read the declaration counts of
\Cref{tab:coverage} from the compiled environment index, which \code{lake exe
library\ub index} prints. The axiom checks of \Cref{sec:eval-axiom-checks} come from running
\code{\#print axioms} on each headline theorem in a clean build, and we cross-check
them with a comment-aware scan for \code{sorry}, \code{admit},
and \code{axiom} over every recorded module and the library closure it imports.

We release the source, the library, and the run record at
\url{https://github.com/Jiyuan-Tan/CausalSmith}, and
\url{https://jiyuan-tan.github.io/CausalSmith/} gives a browsable view of the
library and of the accepted results. The run catalog behind \Cref{tab:funnel,tab:by-cluster}
is the set of banked run directories under \code{CausalSmith/doc/research/\_bank/}, one
directory per run, filed under \code{accepted/}, \code{downgraded/}, or
\code{failed/}. Each directory carries the run's artifacts verbatim, that is, the state
file, the proposal, the review log, and the derivation note, and it is from these that
we read the dispositions and the downgrade reasons we quote. Legacy runs, meaning the
ones that predate the current pipeline, are left out of both tables.

The cluster labels in \Cref{tab:by-cluster} are \CausalSmith{}'s own six clusters. Each
run's state file records its cluster, and the run identifier carries it as a prefix
(\code{stat\_}, \code{exp\_}, \code{panel\_}, \code{eid\_}, \code{pid\_},
\code{scm\_}). Discovery assigns the cluster at proposal time and picks the
cluster-specific setup prompt from it, so the grouping is fixed before anyone knows the
run's mathematics or how it will turn out. The question-type labels of
\Cref{tab:by-type} are a different matter, since the pipeline does not produce them. We
assigned them ourselves, by the rule of \Cref{sec:eval-question-type}, from each run's
proposal-time \code{topic} string. That string is the \code{topic} field in the banked
run's \code{README.md}, and it too is fixed before the run's mathematics, so a reader
who would draw the boundaries elsewhere can apply their own rule to the same text. For
the stated-versus-inferred provenance reported in \Cref{sec:eval-question-type} we read
the anchor paper of each accepted run; we did not repeat that reading for the
downgraded and failed runs.

\paragraph{Base models.}
Every pipeline stage dispatches to one of two agent runners, OpenAI \code{codex} or
Anthropic \code{claude}. Each logical role maps to a model id that is committed in the
pipeline source and can be overridden per role by an environment variable
(\code{CAUSALEAN\_MODEL\_*}). By the close of the campaign the assignment stood as
follows. GPT-5.6 Sol held the hard-mathematics tier, which is the proposal, the
derivation, the referee review, and the \lean{} proof filler, and it also served as the
\code{codex} half of the dual convergence review. GPT-5.6 Terra held the mechanical
tier: scaffold translation, banking, proposal screening, and artifact emission.
GPT-5.5 drafted and revised the presentation. Claude did the planning, the \lean{}
code review, the second convergence review, and the presentation rubric scoring.

Both sides were updated during the campaign. On the OpenAI side the hard-mathematics
tier moved from GPT-5.5 to GPT-5.6 Sol on 2026-07-10, and for the day before that the
proof-filler role sat on the mechanical tier as well. On the Anthropic side the Claude
roles ran as Opus~4.7 until 2026-05-28, as Opus~4.8 until 2026-07-24, and as Opus~5
thereafter. Our campaign ran from 2026-05-14 to 2026-08-26, so cut at those two dates
the $144$ runs of \Cref{tab:funnel} divide into $81$ under Opus~4.7, $45$ under
Opus~4.8, and $18$ under Opus~5, with none spanning a change; of the fourteen accepted
runs, nine sit entirely under Opus~4.8 and the other five entirely under Opus~5.

\end{document}